\newif\ifanonymous
\anonymousfalse

\documentclass[11pt]{article}
\ifanonymous
\PassOptionsToPackage{review}{acl}
\else
\PassOptionsToPackage{preprint}{acl}
\fi
\usepackage{acl}
\usepackage{times}
\usepackage{latexsym}
\usepackage[T1]{fontenc}
\usepackage[utf8]{inputenc}
\usepackage{microtype}
\usepackage{inconsolata}
\usepackage{graphicx}
\usepackage{multirow}
\usepackage{enumitem}
\usepackage{tikz}
\usetikzlibrary{arrows.meta,calc,fit,backgrounds,shadows,shapes.geometric}
\usepackage{flafter}
\usepackage{booktabs}
\usepackage{amsmath}
\usepackage{amssymb}
\usepackage{amsthm}
\usepackage{algorithm}
\usepackage{algpseudocode}
\newtheorem{proposition}{Proposition}
\newtheorem{lemma}[proposition]{Lemma}
\newtheorem{corollary}[proposition]{Corollary}
\usepackage{bm}
\newcommand{\mat}[1]{\bm{#1}}
\renewcommand{\vec}[1]{\bm{#1}}
\usepackage{xcolor}
\usepackage{colortbl}
\newcommand{\numKKfidSteps}{200}  %

\newcommand{\method}{{HiLo}}
\newcommand{\gram}{{Spar}}
\newcommand{\papertitle}{Train High, Ship Low: \\ Efficient LoRA Synchronization for Distributed On-Policy Post-Training}
\ifanonymous
\title{\papertitle}
\author{Anonymous submission}
\else
\title{\raisebox{-0.9em}{\includegraphics[width=3.4em]{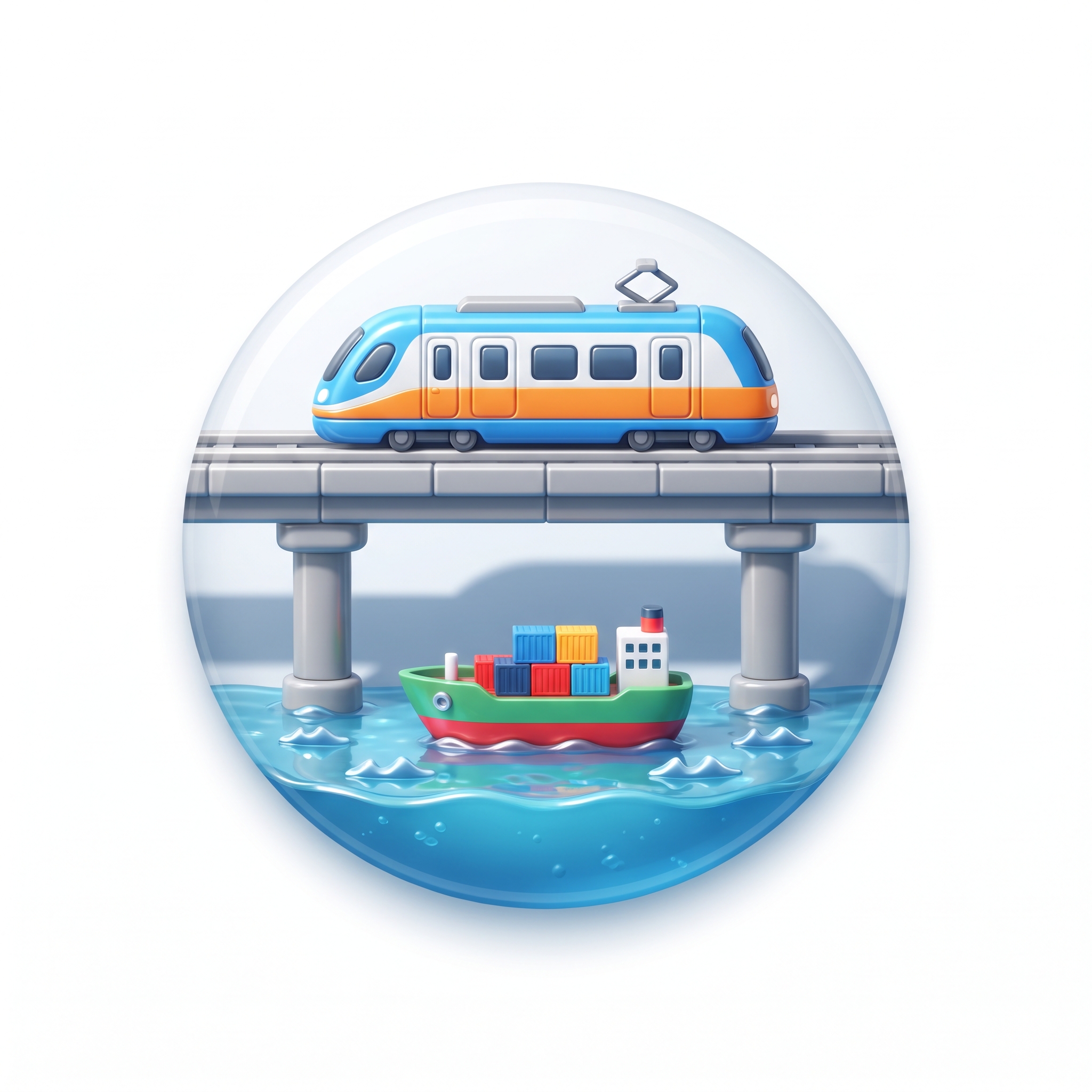}}\hspace{0.25em}\papertitle}
\author{Shenghui Li\textsuperscript{1} \quad Li Ju\textsuperscript{1} \quad Thiemo Voigt\textsuperscript{1,2} \\
{\normalfont\normalsize \textsuperscript{1}Uppsala University, Uppsala, Sweden} \\
{\normalfont\normalsize \textsuperscript{2}Research Institutes of Sweden, Stockholm, Sweden} \\
{\normalfont\small\ttfamily \{shenghui.li,thiemo.voigt\}@angstrom.uu.se \quad li.ju@it.uu.se}}
\fi
\begin{document}
\maketitle

\begin{abstract}
Transmitting LoRA adapters to rollout workers can bottleneck distributed on-policy post-training over bandwidth-constrained networks. Lowering the training rank may limit the capacity a workload needs, while adaptive optimizer updates are difficult to compress. Rollout workers can instead reconstruct updates from gradients by maintaining replicated optimizer states. We introduce \textbf{\method{}}, a compressor-agnostic framework that replaces parameter transmission with compressed gradient dissemination and local optimizer replay. By applying the same decoded gradients, the learner and workers maintain consistent policy states at the same version under reproducible execution. We further develop \textbf{\gram{}}, a budget-aware gradient compressor that combines spectral truncation, adaptive rank allocation and error feedback to decouple training rank from communication cost. Experiments across on-policy distillation and reward-guided post-training show that \method{} with \gram{} maintains learning quality close to full synchronization while transmitting as little as 1.6\% of the adapter bytes, and accelerates end-to-end updates by up to $2.2\times$ under bandwidth constraints.
\end{abstract}

\section{Introduction}

Post-training has become a key stage in improving the capabilities of large language models (LLMs) \citep{ouyang2022instructgpt,rafailov2023dpo,guo2025deepseekr1}.
On-policy methods such as on-policy distillation (OPD) and reinforcement learning with verifiable rewards (RLVR) learn from trajectories sampled from their own evolving policies, guided by teacher supervision or verifier rewards, respectively~\citep{agarwal2024gkd,shao2024deepseekmath}.
The high cost of generating these trajectories motivates practical systems to parallelize rollout generation across many workers \citep{sheng2025hybridflow}. Recent systems such as INTELLECT-2~\citep{primeintellect2025intellect2} and ECHO-2~\citep{song2026echo2} disaggregate policy learning and rollout generation across geographically distributed resources, using Shardcast and pipelined broadcast, respectively, to disseminate policy weights from the learner to the workers over wide-area networks (WANs). Unlike off-policy distillation, on-policy post-training relies on fresh trajectories sampled from the evolving learner policy, so delayed synchronization can introduce rollout--learner mismatch and degrade learning~\citep{kang2026asyncopd}.

Parameter-efficient fine-tuning, particularly Low-Rank Adaptation (LoRA) \citep{hu2022lora}, reduces the synchronization payload to the adapter factors, whose size scales with the training rank. Synchronization can remain expensive over bandwidth-constrained WANs, especially when many workers share the learner's uplink~\citep{song2026echo2}. Although lowering the training rank reduces this cost, it also restricts the space of learnable weight updates. The rank needed for effective adaptation varies across workloads and adapted modules~\citep{zhang2023adaptive}, with very low ranks sufficient in some RL settings but inadequate for supervised fine-tuning or continued pretraining in others~\citep{biderman2024loralearnsless,schulman2025lorawithoutregret}. Ideally, the rank should be determined by the learning workload, while synchronization cost should be controlled independently by the network budget.

Recent work reduces synchronization traffic by losslessly encoding differences between consecutive policy snapshots. SparseRL-Sync transmits changed values and their indices \citep{hu2026sparserlsync},
while ARC exploits unchanged values and bit-level redundancy \citep{fireworks2026arcv3}. These methods preserve the BF16 snapshots exactly and can achieve substantial savings when rounding leaves many weights unchanged. Their payloads, however, depend on the changes that must be preserved and the redundancy available to the
codec. Consequently, lossless encoding alone does not guarantee a prescribed per-update communication budget.

We therefore ask whether synchronization must operate on parameter changes at all. For a gradient-driven optimizer with reproducible transitions, the next state is determined by the current parameters, optimizer state, gradient and update configuration. Distributed data-parallel training uses this principle to keep replicas synchronized through shared gradients and local optimizer updates~\citep{li2020pytorchdistributed}. A rollout worker can likewise reconstruct the learner's update from the same gradient by maintaining replicated parameters and optimizer state under matched execution. This moves compression before the optimizer step, allowing established gradient-compression methods~\citep{lin2018dgc,vogels2019powersgd} to exploit gradient structure before optimizer transformations reshape the update~(\S\ref{sec:optimizer-instances}).

Building on this insight, we introduce \textbf{\method{}} (\textbf{Hi}gh-rank in training, \textbf{Lo}w-rank on the wire),
a framework for communication-efficient LoRA synchronization in
distributed on-policy post-training. \method{} replaces parameter transmission with compressed gradient
dissemination and local optimizer replay, reducing synchronization
traffic without lowering the training rank. We further develop \textbf{Spar} (Spectral Packing with Adaptive Ranks),
a budget-aware gradient compressor tailored to LoRA that serves as
a backend for \method{}. Together, \method{} and Spar alleviate the synchronization bottleneck, allowing distributed rollout to scale more effectively over bandwidth-constrained networks. Our main contributions are as follows.
\begin{itemize}[leftmargin=*,labelsep=0.5em,noitemsep,topsep=3pt]
	\item We formalize \method{} as a compressor-agnostic
	optimizer-replay protocol and establish conditions for matching
	learner and worker parameters and optimizer states at the same
	policy version. This enables different gradient-compression backends to plug into the same synchronization protocol while preserving same-version consistency.
	
	\item We design Spar to compress LoRA-factor gradients using
	spectral truncation via small Gram-matrix eigendecompositions,
	global rank allocation under a byte budget, and error feedback. The resulting allocation prioritizes gradient directions with the highest spectral value per communicated byte.
	
	\item We evaluate \method{} on two reasoning tasks and test
	robustness across four optimizers. With Spar, \method{} maintains
	accuracy close to full synchronization at 1.6\% of the adapter
	bytes and achieves up to $2.2\times$ end-to-end speedup on
	bandwidth-limited links.
\end{itemize}

\begin{figure*}[!t]
\centering
\input{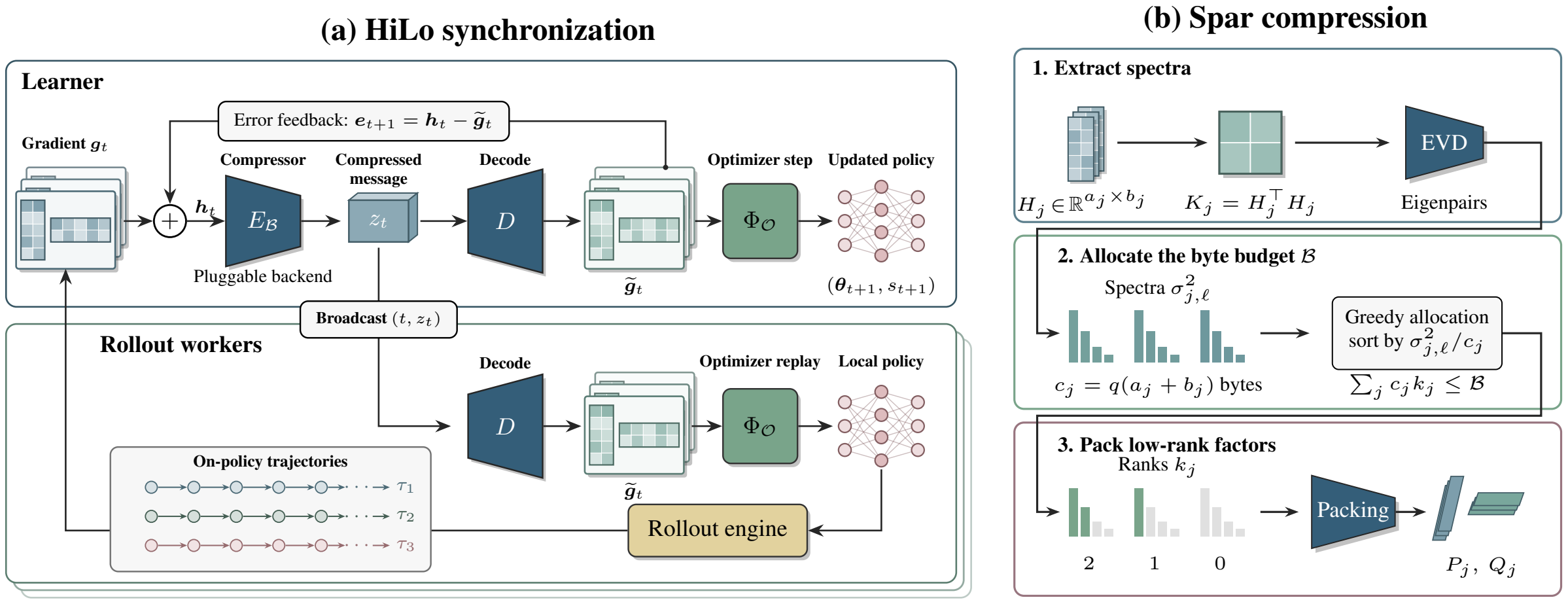}
\caption{\textbf{HiLo with the Spar compression backend.} (a) The learner compresses error-corrected LoRA-factor gradients and broadcasts $z_t$. The learner and rollout workers decode the same gradient and apply the same optimizer transition, maintaining matching parameters and optimizer states under the conditions in \S\ref{sec:replay}. (b) Spar extracts spectra through small Gram eigendecompositions, greedily allocates communication ranks under a global factor-payload budget, and packs low-rank factors for transmission. Spectra and ranks are illustrative; the Gram construction shown is for $a_j\geq b_j$, with the other case symmetric.}
\label{fig:architecture}
\label{fig:spar}
\end{figure*}

\section{Background and Preliminaries}
\label{sec:motivation}
\subsection{On-Policy Post-Training}
On-policy post-training improves a student policy $\pi_{\vec{\theta}}$ by learning from responses $y\sim\pi_{\vec{\theta}}(\cdot\mid x)$ sampled from its own evolving distribution. Two representative approaches are reinforcement learning with verifiable rewards (RLVR) and on-policy distillation (OPD). In RLVR, a verifier evaluates responses using task-specific correctness signals, and policy-gradient algorithms such as GRPO optimize the student using reward-derived advantages \citep{shao2024deepseekmath}. In OPD, a teacher provides token-level supervision along student-generated trajectories, typically through a divergence-based distillation objective \citep{agarwal2024gkd}. The two signals can also be combined in a single objective. Reinforcement-aware distillation (RLAD) \citep{zhang2026rlad} keeps a reward-driven PPO-style update but anchors its importance ratio partly to the teacher, so the teacher shapes the update without a separate KL term. Their supervision differs, but all of them learn from an evolving student policy.

\subsection{Distributed Rollout}

In actor--learner architectures, rollout workers generate trajectories using local policy copies, while a learner computes gradients and updates parameters~\citep{sheng2025hybridflow}. Trajectories flow from workers to the learner, and after every update the new policy must flow back to every worker. Synchronous execution places this transfer on the critical path and blocks rollout generation until it completes, whereas asynchronous execution overlaps communication with computation at the cost of potentially stale policies. Decentralized systems take the asynchronous route and bound the lag explicitly. INTELLECT-2 lets workers generate with a policy a few updates old while checkpoints propagate through relay servers~\citep{primeintellect2025intellect2}, and ECHO-2 caps staleness while pipelining the broadcast among peers~\citep{song2026echo2}. Both change how and when the weights arrive, but every worker still receives the full policy. Many on-policy objectives, such as distillation, need samples from the current student and tolerate little lag. For instance, synchronizing the full policy only every few updates can noticeably degrade the student once the lag grows (Appendix~\ref{app:syncK}). We study this learner-to-worker path.

\subsection{The Synchronization Bottleneck}
\label{sec:bottleneck}
Let $S$ be the bytes the learner sends per update and $N$ the number of workers. Under unicast over a learner uplink of bandwidth $B$, each update spends about $NS/B$ seconds in transfer, and relay or pipelined broadcast~\citep{primeintellect2025intellect2,song2026echo2} only moves this bottleneck to each worker's downlink. For LoRA, $S$ grows linearly with the training rank $r$. The rank-64 adapter of Qwen3-1.7B is 139.5\,MB in BF16, and at 62.5\,Mbps per worker its transfer alone exceeds the 15\,s of computation per update (\S\ref{sec:systems}).

Existing remedies each give something up. Lowering $r$ caps what the adapter can learn, and the rank a workload needs is not known in advance~\citep{schulman2025lorawithoutregret,biderman2024loralearnsless}. Lossless delta codecs~\citep{fireworks2026arcv3,hu2026sparserlsync} rely on the sparsity of full-parameter updates, whereas a LoRA update is dense. Lossy compression of the adapter or of its change truncates the optimizer's output, which is generically full rank in factor space, and if only the workers adopt the truncated weights, they sample from a policy the learner never held. This same-version discrepancy, $\vec{\theta}^{(i)}_v\ne\vec{\theta}_v$ for worker $i$ at version $v$, is distinct from temporal staleness, the lag between $v$ and the learner's current version. We therefore seek a message whose size is set by a byte budget rather than by $r$, that learner and workers decode to the same state, and that applies to any gradient-driven optimizer. A worker that holds the parameters and optimizer state needs only the gradient, the one input that compresses well, so \method{} sends it and leaves staleness to the schedule.

\section{Proposed Method}
\label{sec:method}
We introduce \textbf{HiLo}, a communication-efficient LoRA synchronization framework that replaces parameter transmission with compressed gradient dissemination and local optimizer replay. As illustrated in Figure~\ref{fig:architecture}, the learner compresses aggregated LoRA-factor gradients with error feedback, while both the learner and rollout workers apply the same decoded gradients through locally maintained optimizer states. \method{} supports different gradient compressors. We further develop \textbf{\gram{}}, a LoRA-specific backend that combines Gram-matrix-based low-rank approximation with budget-aware rank allocation across factors, so that synchronization traffic no longer depends on the training rank.

\subsection{Replay-based LoRA Synchronization}
\label{sec:replay}
Let $\vec{\theta}_t$ collect all trainable LoRA parameters at version $t$. Let $\mathcal O$ be the optimizer, $s_t$ all state required by its update rule, and $\xi_t$ its shared update configuration, including learning rate, hyperparameters and any reproducible stochastic inputs. The framework is defined for transitions fully determined by these inputs and the decoded gradient. The learner and rollout workers are initialized from shared parameters and optimizer state $(\vec{\theta}_0,s_0)$. Each worker subsequently reconstructs policy versions by replaying the corresponding gradient messages in order. Let $\vec{g}_t$ denote the final LoRA-factor gradient after any required aggregation and gradient clipping, but before the optimizer update. The learner maintains an error-feedback residual $\vec{e}_t$ and computes the compressed gradient message as follows:
\begin{equation}
\begin{aligned}
 \vec{h}_t &= \vec{g}_t+\vec{e}_t, &\qquad
 z_t &= E_{\mathcal B}(\vec{h}_t),\\
 \widetilde{\vec{g}}_t &= D(z_t), &\qquad
 \vec{e}_{t+1} &= \vec{h}_t-\widetilde{\vec{g}}_t.
\end{aligned}
\label{eq:ef}
\end{equation}
$E_{\mathcal B}$ encodes under byte budget $\mathcal B$, and $D$ decodes the actual wire message. Thus $\mathcal C_{\mathcal B}=D\circ E_{\mathcal B}$ includes quantization and reconstruction, not just an ideal truncated matrix. The residual need not be replicated because only the learner constructs subsequent compressed gradients. \method{} is agnostic to the compression backend, so any compressor whose decoded gradient $D(z_t)$ is reproducible on every worker can be integrated, including PowerSGD, Top-$k$ sparsification and the budget-aware \gram{} compressor of \S\ref{sec:compression}.

The optimizer transition is
\begin{equation}
(\vec{\theta}_{t+1},s_{t+1})=\Phi_{\mathcal O}(\vec{\theta}_t,s_t,\widetilde{\vec{g}}_t;\xi_t),
\label{eq:replay}
\end{equation}
and worker $i$ replays the same rule on its own copy $(\vec{\theta}^{(i)}_t,s^{(i)}_t)$ with the same decoded gradient $\widetilde{\vec{g}}_t$ and configuration $\xi_t$.
No worker backward pass or teacher query is needed. The protocol covers gradient-driven transitions with replicable state; an optimizer that needs untransmitted activations, objective evaluations, curvature or fresh randomness must have those inputs reproduced or communicated, at their own cost.

\paragraph{Same-version consistency.}
\label{sec:replay-consistency}
\method{} keeps every worker consistent with the learner at the same version, the second requirement of \S\ref{sec:bottleneck}. Both sides start from the same parameters and optimizer state $(\vec{\theta}_0,s_0)$ and apply the same transition $\Phi_{\mathcal O}$ of Equation~\ref{eq:replay} to the same decoded gradient $\widetilde{\vec{g}}_t$, so by induction $(\vec{\theta}^{(i)}_t,s^{(i)}_t)=(\vec{\theta}_t,s_t)$ at every version $t$ that worker $i$ consumes. This requires both sides to decode the same wire message and to replay versioned messages exactly once and in order, and bitwise equality further requires matching numerical execution, optimizer implementation and state precision. Gradient compression therefore changes the shared learning trajectory but adds no learner--worker discrepancy. A lagging worker still reconstructs the learner's earlier state exactly, so \method{} does not remove temporal policy staleness.

\paragraph{Optimizer scope.}
Common optimizers fall within this replicable scope. AdamW requires replicated first and second moments, step counters and matching bias-correction and weight-decay semantics \citep{loshchilov2019adamw}. Muon requires replicated momentum and the same orthogonalization and scaling rules \citep{jordan2024muon}. Lion \citep{chen2023lion} and SGD with momentum are compatible under the same conditions, and \S\ref{sec:cross-optimizer} evaluates all four. Detailed optimizer transitions and initialization are given in Appendix~\ref{app:optimizer-implementation}.

\subsection{Spectral Packing with Adaptive Ranks}
\label{sec:compression}
\gram{} compresses the error-corrected LoRA-factor gradients through spectral truncation and global rank allocation. Let $\mat{H}_j\in\mathbb R^{a_j\times b_j}$ denote the block of $\vec{h}_t$ corresponding to LoRA factor $j$, with one dimension equal to the training rank $r$. For $a_j\ge b_j$, let $\mat{V}_{j,k_j}$ contain the leading $k_j$ eigenvectors of the small Gram matrix $\mat{H}_j^\top\mat{H}_j$. \gram{} encodes the block as
\begin{equation*}
\mat{P}_j=\mat{H}_j\mat{V}_{j,k_j},\qquad
\mat{Q}_j=\mat{V}_{j,k_j}.
\end{equation*}
In exact arithmetic, $\widehat{\mat{H}}_j=\mat{P}_j\mat{Q}_j^\top$ is a best rank-$k_j$ approximation of $\mat{H}_j$ in Frobenius norm. The case $a_j<b_j$ is handled symmetrically (Appendix~\ref{app:evd}). Each retained direction requires $a_j+b_j$ communicated scalars. The transmitted factors are decoded into the gradient used for optimizer replay and residual computation.

Let $\sigma_{j,\ell}$ denote the singular values of $\mat{H}_j$ in descending order and $q$ the bytes per encoded scalar. Each retained direction preserves spectral energy $\sigma_{j,\ell}^2$ at a cost of $c_j=q(a_j+b_j)$ bytes. Under a global factor-payload budget $\mathcal B$, we seek
\begin{equation}
\begin{aligned}
\max_{\{k_j\}}\quad &\sum_j\sum_{\ell=1}^{k_j}\sigma_{j,\ell}^2\\
\text{s.t.}\quad &\sum_j c_j k_j\le\mathcal B,\\
&k_j\in\{0,\ldots,\min(a_j,b_j)\}.
\end{aligned}
\label{eq:budget}
\end{equation}
This objective minimizes the total squared truncation error before wire encoding. We use a greedy heuristic that sorts directions by $\sigma_{j,\ell}^2/c_j$ and retains the longest prefix whose cumulative cost fits $\mathcal B$ (Appendix~\ref{app:evd}). The rule favors spectral energy per byte but is not guaranteed to be globally optimal when costs differ. The allocated ranks vary across factors and updates while the training rank $r$ remains fixed. For comparison, a uniform communication rank $k$ defines the reference budget
\begin{equation}
\mathcal B_k=qk\sum_j(a_j+b_j).
\label{eq:budget-k}
\end{equation}

\subsection{Why the Gradient Rather than the Update}
\label{sec:optimizer-instances}
Compressing after the optimizer, as the parameter-side methods of \S\ref{sec:bottleneck} do (the Delta baseline of \S\ref{sec:setup} and LoGRA), truncates a different object. Under SGD the update is a multiple of the gradient and both truncations discard the same energy, but momentum raises the update's rank by the communication rank at every step, and AdamW's coordinatewise normalization replaces the gradient's spectrum with that of its second-moment statistics, so a rank-one gradient stream generically yields a full-rank update in factor space (Appendix~\ref{app:rank}). In a synthetic AdamW replay, a rank-one truncation keeps all of the gradient but only 20--40\% of the update (Table~\ref{tab:rank-witness}).

\paragraph{Worker cost.}
Replay costs worker memory and computation. For $|\vec{\theta}|$ trainable LoRA entries, FP32 AdamW moments cost $8|\vec{\theta}|$ bytes and FP32 Muon momentum $4|\vec{\theta}|$, plus master parameters and buffers where used, and replay adds the optimizer's local update cost, including orthogonalization for Muon. For a factor of shape $a\times b$ decoded at rank $k$, reconstruction costs $O(abk)$ operations. Encoder timings are in \S\ref{sec:systems} and the cost model in Appendix~\ref{app:model}.

\section{Experiments}
\label{sec:evaluation}
\begin{table*}[!t]
\centering
\footnotesize
\setlength{\tabcolsep}{3.4pt}
\renewcommand{\arraystretch}{1.18}
\providecommand{\std}[1]{\,{\color{black!72}\scriptsize$\pm$#1}}
\definecolor{oursrow}{RGB}{232,240,252}
\definecolor{taskrow}{gray}{0.91}
\definecolor{headrow}{gray}{0.96}
\begin{tabular}{l cccc c cccc}
\toprule
& \multicolumn{4}{c}{\textbf{pass@4 (\%)} $\uparrow$} & & \multicolumn{4}{c}{\textbf{mean@4 (\%)} $\uparrow$} \\
\cmidrule(lr){2-5}\cmidrule(l){7-10}
\rowcolor{taskrow}\multicolumn{10}{l}{\textbf{Knights and Knaves} \ {\color{black!70}$\cdot$\ OPD\ $\cdot$\ Qwen3-1.7B}} \\
\rowcolor{headrow}{\color{black!70}Rank $k$ (payload)} & $1$\,{\color{black!70}(1.6\%)} & $2$\,{\color{black!70}(3.3\%)} & $4$\,{\color{black!70}(6.6\%)} & $8$\,{\color{black!70}(13.1\%)} & & $1$\,{\color{black!70}(1.6\%)} & $2$\,{\color{black!70}(3.3\%)} & $4$\,{\color{black!70}(6.6\%)} & $8$\,{\color{black!70}(13.1\%)} \\
\cmidrule(lr){2-5}\cmidrule(l){7-10}
{\color{black!60}Full Sync} & \multicolumn{4}{c}{{\color{black!60}83.2\std{1.2}}} & & \multicolumn{4}{c}{{\color{black!60}48.6\std{1.0}}} \\
Delta & 75.4\std{1.1} & 81.8\std{2.3} & 80.9\std{1.5} & 81.8\std{2.4} & & 40.6\std{1.8} & 46.7\std{3.0} & 44.7\std{0.9} & 46.0\std{3.0} \\
Low-rank LoRA ($r{=}k$) & 78.0\std{2.4} & 79.6\std{3.0} & 81.0\std{0.9} & 82.0\std{1.0} & & 42.8\std{1.8} & 43.6\std{1.7} & 45.3\std{1.1} & 47.3\std{0.3} \\
\method{} + PowerSGD & 78.8\std{1.5} & 81.7\std{0.7} & 80.5\std{2.7} & 81.5\std{1.3} & & 42.9\std{1.1} & 46.6\std{0.9} & 45.6\std{2.0} & \textbf{47.8}\std{1.9} \\
\method{} + Top-$k$ & 76.6\std{3.1} & 79.0\std{2.3} & 81.4\std{2.0} & 81.1\std{0.7} & & 41.3\std{2.2} & 43.2\std{1.9} & 45.6\std{2.2} & 46.5\std{0.2} \\
\rowcolor{oursrow}\textbf{\method{} + \gram{} (ours)} & \textbf{81.6}\std{1.8} & \textbf{83.1}\std{1.2} & \textbf{81.8}\std{2.0} & \textbf{82.3}\std{0.8} & & \textbf{45.7}\std{1.6} & \textbf{47.5}\std{1.2} & \textbf{46.8}\std{2.1} & 47.5\std{0.8} \\
\midrule
\rowcolor{taskrow}\multicolumn{10}{l}{\textbf{Countdown-4} \ {\color{black!70}$\cdot$\ RLAD\ $\cdot$\ Qwen3-0.6B}} \\
\rowcolor{headrow}{\color{black!70}Rank $k$ (payload)} & $0.5$\,{\color{black!70}(0.78\%)} & $1$\,{\color{black!70}(1.6\%)} & $2$\,{\color{black!70}(3.1\%)} & $4$\,{\color{black!70}(6.2\%)} & & $0.5$\,{\color{black!70}(0.78\%)} & $1$\,{\color{black!70}(1.6\%)} & $2$\,{\color{black!70}(3.1\%)} & $4$\,{\color{black!70}(6.2\%)} \\
\cmidrule(lr){2-5}\cmidrule(l){7-10}
{\color{black!60}Full Sync} & \multicolumn{4}{c}{{\color{black!60}64.0\std{0.7}}} & & \multicolumn{4}{c}{{\color{black!60}50.7\std{1.2}}} \\
Delta & -- & 59.7\std{2.8} & 62.9\std{1.6} & 63.3\std{1.1} & & -- & 46.8\std{2.9} & 49.0\std{2.0} & 50.7\std{2.0} \\
Low-rank LoRA ($r{=}k$) & -- & 61.0\std{0.6} & 62.7\std{0.7} & 63.9\std{0.5} & & -- & 47.3\std{0.5} & 48.6\std{1.1} & 49.8\std{0.7} \\
\method{} + PowerSGD & -- & 61.2\std{0.8} & 63.1\std{0.1} & 63.3\std{1.5} & & -- & 48.2\std{0.6} & \textbf{51.1}\std{0.8} & 51.5\std{2.2} \\
\method{} + Top-$k$ & 57.9\std{1.3} & 60.6\std{0.4} & 62.1\std{0.7} & 63.8\std{0.2} & & 44.5\std{0.9} & 47.6\std{0.2} & 49.0\std{0.5} & 50.4\std{1.2} \\
\rowcolor{oursrow}\textbf{\method{} + \gram{} (ours)} & \textbf{62.5}\std{1.9} & \textbf{63.7}\std{1.5} & \textbf{63.9}\std{0.4} & \textbf{64.7}\std{0.7} & & \textbf{49.5}\std{2.4} & \textbf{51.0}\std{0.7} & 51.0\std{0.6} & \textbf{52.2}\std{0.9} \\
\bottomrule
\end{tabular}
\caption{Held-out accuracy (\%) on KK (update 300) and Countdown-4 (update 200), mean $\pm$ s.d.\ over three seeds. Payload is relative to fully synchronizing the rank-64 adapter. Low-rank LoRA reduces the training rank to $k$, whereas all \method{} variants keep rank 64 and compress only the communicated gradients; a fractional $k$ denotes a global byte budget. Delta uses error feedback to compress post-optimizer changes at the same payload; the learner adopts the compressed weights, so workers keep no optimizer state.
}
\label{tab:main_accuracy}
\end{table*}

In this section, we evaluate \method{} on two reasoning tasks. \S\ref{sec:setup} describes the setup and verifies that replay reproduces the learner's state, \S\ref{sec:main-results} compares learning quality against full synchronization and the baselines at matched payloads, and \S\ref{sec:systems} measures payload, overhead and end-to-end step time with live worker replay.

\subsection{Experimental Setup}
\label{sec:setup}

\paragraph{Models and datasets.}
We evaluate on two reasoning tasks from Reasoning Gym \citep{stojanovski2025reasoninggym} that differ in their training signal. \emph{Knights and Knaves} (KK) \citep{xie2024memorization} asks which of six inhabitants tell the truth and which lie, given their statements. We train it with pure OPD, minimizing the sampled-token reverse KL divergence to a Qwen3-8B teacher \citep{qwen2025qwen3}. \emph{Countdown-4} \citep{gandhi2024sos} asks for an arithmetic expression over four given numbers that reaches a target value. We train it with the RLAD objective \citep{zhang2026rlad}, combining a verifiable task reward with the log-probabilities of the same teacher (Appendices~\ref{sec:countdown} and~\ref{app:rlad}). The main KK student is Qwen3-1.7B-Base with a rank-64 LoRA adapter, and the Countdown-4 student is Qwen3-0.6B-Base with a rank-64 LoRA adapter. All adapters use the rank-stabilized scale $s=\alpha/\sqrt{r}$ for LoRA \citep{kalajdzievski2023rslora}. For both tasks, we report accuracy on 1,000 held-out problems disjoint from the training set, with a 2,048-token generation cap.

\paragraph{Baselines.}
\textbf{Full Sync} trains a rank-64 adapter and ships it in full after every update, the uncompressed reference. \textbf{Low-rank LoRA ($r{=}k$)} lowers the training rank to $k$ and fully synchronizes the adapter. \textbf{Delta} keeps rank 64 and instead compresses the post-optimizer parameter change. The learner adds its carried error, ships the rank-$k$ truncation and adopts the truncated weights itself, so workers need no optimizer state. \method{} keeps rank 64 and compresses only the gradient. Besides our \textbf{\gram{}} backend, we instantiate it with \textbf{PowerSGD} \citep{vogels2019powersgd} using one warm-started power iteration and with \textbf{Top-$k$} \citep{lin2018dgc} sending a BF16 value and an int32 index per entry. All \method{} variants apply error feedback \citep{karimireddy2019ef} before the optimizer and are matched to the analytical rank-$k$ payload $\mathcal B_k$ of Equation~\ref{eq:budget-k}, so they differ only in the backend (Appendix~\ref{app:baselines}).

\paragraph{Implementation details.}
For \gram{}, the low-rank factors are transmitted in BF16 ($q=2$ bytes per scalar), with error feedback computed against the decoded gradients.
Unless stated otherwise, all methods use AdamW with a learning rate of $3\times10^{-4}$ with linear decay, and every configuration is run with three seeds. Each KK update samples 16 prompts $\times$ 8 rollouts, and KK training runs for 300 updates. Each Countdown-4 update samples 64 prompts $\times$ 8 rollouts, keeps the 32 groups with non-constant reward, and the schedule spans 200 updates after a 10-step warmup. KK actors serve the LoRA adapter unmerged through vLLM's LoRA kernels \citep{kwon2023vllm}, whereas Countdown-4 actors serve merged BF16 weights and may generate one update behind the learner. In every compressed method the learner itself follows the compressed trajectory, so actors never sample from parameters the learner does not hold. For the \method{} variants, actors obtain their parameters through a replay simulator that applies each decoded gradient to a replicated optimizer state, and its output matches the learner bitwise in every run we checked; for Delta they receive the truncated weights (Appendix~\ref{app:baselines}). All runs use NVIDIA GH200 GPUs. At the end of training, we evaluate four samples per problem at temperature 1.0 and report pass@4 and mean@4.

\paragraph{Replay consistency check.}
\label{sec:replay-check}
Before comparing methods, we verify the same-version consistency of optimizer replay (\S\ref{sec:replay-consistency}). A Qwen3-0.6B-Base learner with rank-32 LoRA and \gram{} at rank 1 consumes exactly the decoded BF16 gradient and records the initial LoRA parameters and every shipped factor. A separate process on another GH200 GPU replays AdamW from these inputs alone. Over 60 KK updates, all 392 LoRA tensors match the learner bitwise at every tenth update (maximum absolute difference 0). Both processes share the GPU type and software stack, so heterogeneous execution remains unverified (Appendix~\ref{app:replay-check}).

\subsection{Learning Quality under Compression}
\label{sec:main-results}

\begin{figure}[t]
\centering
\includegraphics[width=\linewidth]{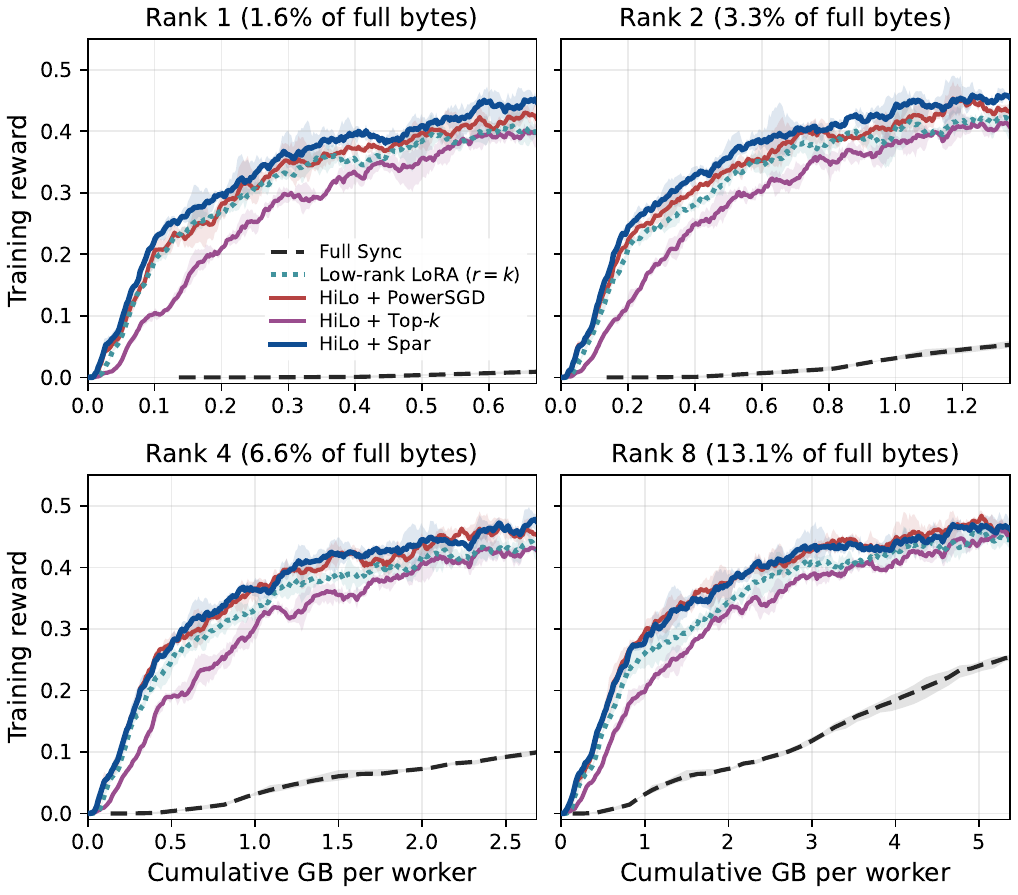}
\caption{Training reward against cumulative payload per worker for KK OPD with Qwen3-1.7B-Base over the first 300 updates. Each panel fixes the communication rank $k$; the x-axis ends at the compressed methods' 300-update total. Full Sync ships 139.5\,MB per update.}
\label{fig:kk17_bytes_full}
\end{figure}

\begin{figure}[t]
\centering
\includegraphics[width=\linewidth]{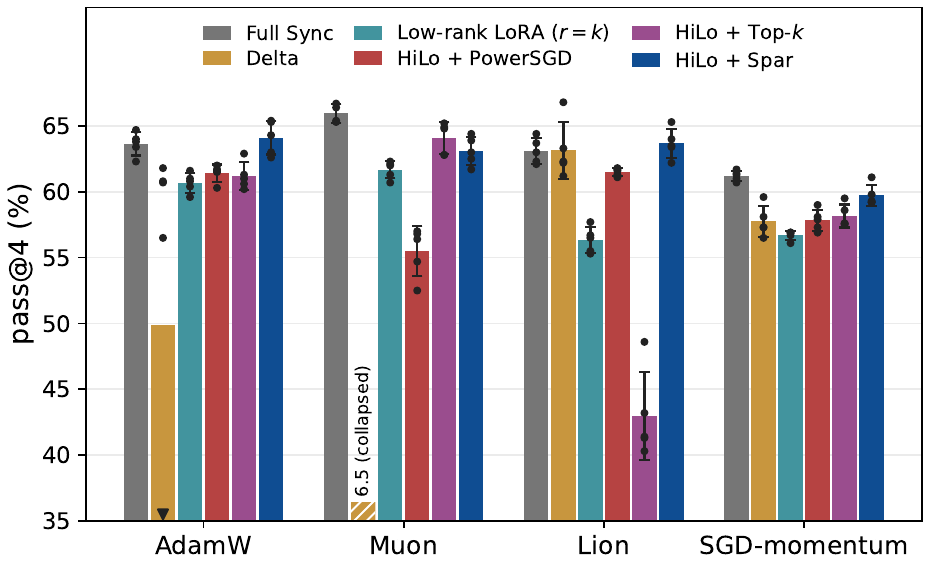}
\caption{Optimizer robustness on Countdown-4 at $k=1$. \method{} + \gram{} is the only compressed method that stays within 3 pass@4 points of Full Sync across all four optimizers, while competing methods suffer substantial degradation under certain optimizers. }
\label{fig:countdown_optimizers}
\end{figure}

\paragraph{Accuracy at matched payloads.}
Table~\ref{tab:main_accuracy} shows that keeping the training rank and compressing only the gradient preserves learning quality, with the largest benefit from \gram{} at the tightest budgets. On KK at rank 1 (1.6\% of the bytes), \method{} + \gram{} reaches 81.6\% pass@4 against 83.2\% for Full Sync, ahead of \method{} + PowerSGD (78.8\%), Low-rank LoRA (78.0\%) and \method{} + Top-$k$ (76.6\%), and it matches Full Sync at rank 2. Delta trails it by 6.2 points at rank 1, because the gradient's rank-$k$ tail is set by the data whereas the AdamW update's is set by the second-moment statistics (\S\ref{sec:optimizer-instances}). On Countdown-4 under RLAD, \gram{} reaches 63.7\% at rank 1 against 64.0\% for Full Sync, ahead of PowerSGD and Low-rank LoRA in every seed, and its global budget still reaches 62.5\% at 0.78\% of the bytes. Low-rank LoRA trails \method{} + \gram{} at the smallest ranks on both tasks, which supports keeping the training rank. As the budget grows the backends converge, so the backend matters most when bandwidth is scarce. With three seeds, the margin over PowerSGD at a single rank is within about two standard deviations, and the evidence is the consistent trend across ranks and tasks.

\paragraph{Communication efficiency.}
Figure~\ref{fig:kk17_bytes_full} shows the same comparison against communication on KK. At ranks 1, 2 and 4, \method{} + \gram{} reaches a given training reward with the fewest cumulative bytes, and at rank 8 it is on par with PowerSGD. Full Sync is still near its initial reward when the compressed methods have finished 300 updates.

\paragraph{Ablation and optimizer robustness.}
\label{sec:cross-optimizer}
Table~\ref{tab:spar_ablation} removes one component at a time on Countdown-4 at the same payload. At $k{=}1$ the global budget carries the gain. Exact truncation with one direction per factor only matches PowerSGD (61.4\% pass@4), while \gram{} reaches 64.1\%, ahead in four of five seeds. At $k{=}2$ the variants with error feedback are within one point; without it, accuracy drops by 1.9 points at $k{=}2$ and two of five seeds collapse at $k{=}1$.
\begin{table}[t]
\centering
\footnotesize
\setlength{\tabcolsep}{3.4pt}
\renewcommand{\arraystretch}{1.18}
\providecommand{\std}[1]{\,{\color{black!72}\scriptsize$\pm$#1}}
\definecolor{oursrow}{RGB}{232,240,252}
\definecolor{headrow}{gray}{0.96}
\begin{tabular}{l cc}
\toprule
& \multicolumn{2}{c}{\textbf{pass@4 (\%)} $\uparrow$} \\
\cmidrule(l){2-3}
\rowcolor{headrow}{\color{black!70}Rank $k$ (payload)} & $1$\,{\color{black!70}(1.6\%)} & $2$\,{\color{black!70}(3.1\%)} \\
\midrule
{\color{black!60}Full Sync} & \multicolumn{2}{c}{{\color{black!60}63.6\std{0.9}}} \\
Exact, $k$ per factor & 61.4\std{2.1} & \textbf{64.1}\std{0.8} \\
PowerSGD ($k$ per factor) & 61.4\std{0.7} & 63.1\std{0.1} \\
\gram{} w/o EF & 62.1\std{0.8}$^{2/5}$ & 62.0\std{1.7} \\
\rowcolor{oursrow}\textbf{\gram{} (ours)} & \textbf{64.1}\std{1.3} & 63.9\std{0.4} \\
\bottomrule
\end{tabular}
\caption{\gram{} ablation on Countdown-4 at matched payload (AdamW; 5 seeds at $k{=}1$, 3 at $k{=}2$). $^{c/n}$: $c$ of $n$ seeds collapsed and are excluded. Bold marks the best compressed variant per column.}
\label{tab:spar_ablation}
\end{table}

Because \method{} replays whatever transition the optimizer defines, we repeat the Countdown-4 comparison at rank 1 under AdamW, Muon \citep{jordan2024muon}, Lion \citep{chen2023lion} and SGD with momentum, with five seeds per arm and the learning rate tuned for Full Sync per optimizer (Appendix~\ref{sec:countdown}). Figure~\ref{fig:countdown_optimizers} shows that \method{} + \gram{} is the only compressed method that stays within 3 pass@4 points of Full Sync under all four optimizers. PowerSGD drops to 55.5\% under Muon, Top-$k$ drops to 43.0\% under Lion, and Low-rank LoRA trails \gram{} under every optimizer. Delta is the least robust, collapsing in all five Muon seeds and in one of five AdamW seeds, whereas no \method{} + \gram{} run collapses. Delta's ordering across optimizers follows \S\ref{sec:optimizer-instances}. Momentum raises the update's rank only additively, AdamW makes it generically full, and Muon's orthogonalization also flattens its spectrum, which makes truncation least effective. Delta's parity with \method{} + \gram{} under Lion is not predicted by this rank argument (Appendix~\ref{app:rank-optimizer}).

\subsection{System Evaluation}
\label{sec:systems}

\begin{figure}[t]
\centering
\includegraphics[width=\linewidth]{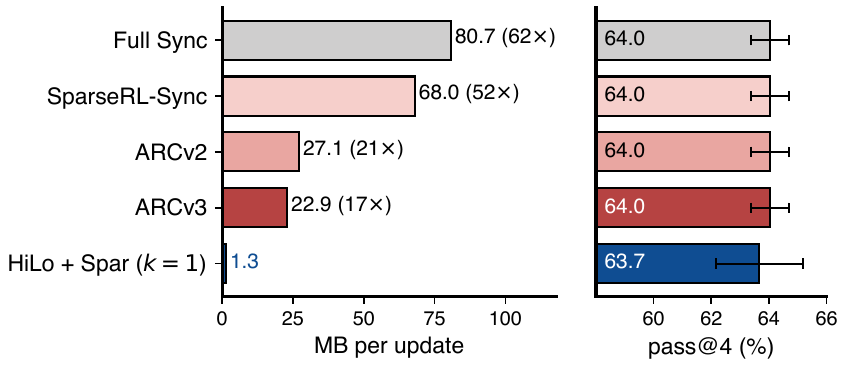}
\caption{Per-update communication payload and pass@4 on Countdown-4. \method{} + \gram{} ($k=1$) sends 62$\times$ fewer bytes than full adapter synchronization and 17$\times$ fewer than the strongest lossless delta codec, at comparable accuracy. Payload ratios are relative to \method{}; the lossless methods reproduce Full Sync exactly and share its accuracy.}
\label{fig:sync_bars}
\end{figure}

\paragraph{Comparison with lossless weight synchronization.}
Deployed systems shrink the policy payload by encoding the change between consecutive weight versions without loss \citep{fireworks2026arcv3,hu2026sparserlsync}. On the rank-64 Countdown-4 adapter (Figure~\ref{fig:sync_bars}), about 61\% of the BF16 elements change between consecutive updates, so SparseRL-Sync, which sends the changed values and their indices, saves only 16\% over Full Sync. ARCv2 and ARCv3 reduce the payload to 27.1 and 22.9\,MB, whereas \method{} + \gram{} at $k{=}1$ sends 1.3\,MB, 17$\times$ less, at a cost of 0.3 pass@4 points. ARCv3 also takes 12--14\,s to encode each update.

\paragraph{Encoder and replay overhead.}
On one GH200, the \gram{} encoder takes 25--27\,ms per update for the 392 factor gradients of the rank-64 Qwen3-1.7B adapter, against 35--41\,ms for one-iteration PowerSGD, 22--24\,ms for Top-$k$ and about 550\,ms for exact batched SVD, and a worker rebuilds the gradient in 13--19\,ms. \gram{}, PowerSGD and Top-$k$ thus stay below 0.5\% of a 15\,s lockstep update (Appendix~\ref{app:model}). The end-to-end runs below used the slower \texttt{torch.linalg.eigh} path, about 0.27\,s more per update (Appendix~\ref{app:evd}).

\paragraph{End-to-end step time.}
We run the Qwen3-1.7B KK recipe with one learner GPU and eight workers on three GH200 nodes (Table~\ref{tab:e2e}). Each worker scores its own rollouts with the Qwen3-8B teacher, and under \method{} it receives the BF16 rank-1 gradient factors and replays AdamW live. Links are TCP-paced to a fixed bandwidth per worker, and 62.5 and 125\,Mbps per worker correspond to unicast learner uplinks of 0.5 and 1\,Gbps, the same order as the WAN regimes of ECHO-2 \citep{song2026echo2}.

\begin{table}[t]
\centering
\footnotesize
\setlength{\tabcolsep}{3.4pt}
\renewcommand{\arraystretch}{1.18}
\providecommand{\std}[1]{\,{\color{black!72}\scriptsize$\pm$#1}}
\definecolor{oursrow}{RGB}{232,240,252}
\definecolor{headrow}{gray}{0.96}
\resizebox{\linewidth}{!}{%
\begin{tabular}{l ccc c ccc}
\toprule
& \multicolumn{3}{c}{\textbf{Lockstep (lag 0)}: s/update $\downarrow$} & & \multicolumn{3}{c}{\textbf{One-step-off (lag 1)}: s/update $\downarrow$} \\
\cmidrule(lr){2-4}\cmidrule(l){6-8}
\rowcolor{headrow}{\color{black!70}Mbps/worker} & {\color{black!70}Full Sync} & \textbf{\method{}} & {\color{black!70}Speedup} & & {\color{black!70}Full Sync} & \textbf{\method{}} & {\color{black!70}Speedup} \\
\midrule
unpaced & {\color{black!60}15.3\std{0.1}} & 15.9\std{0.1} & 0.97$\times$ & & -- & -- & -- \\
$125$ & {\color{black!60}24.2\std{0.1}} & \textbf{16.2}\std{0.3} & \textbf{1.50$\times$} & & {\color{black!60}11.9\std{0.0}} & \textbf{8.1}\std{0.0} & \textbf{1.48$\times$} \\
$62.5$ & {\color{black!60}33.1\std{0.1}} & \textbf{16.1}\std{0.2} & \textbf{2.06$\times$} & & {\color{black!60}17.9\std{0.0}} & \textbf{8.1}\std{0.1} & \textbf{2.22$\times$} \\
\bottomrule
\end{tabular}}
\caption{Measured seconds per update for KK OPD with Qwen3-1.7B-Base, one learner and eight rollout workers with the teacher. Full synchronization ships the 139.5\,MB adapter while \method{} ships 2.3\,MB of rank-1 gradient factors.}
\label{tab:e2e}
\end{table}

In lockstep, an update costs about 15\,s of computation. Full synchronization adds the transfer of the 139.5\,MB adapter, reaching 24.2\,s at 125\,Mbps and 33.1\,s at 62.5\,Mbps, whereas \method{} stays near 16\,s, a $1.5\times$ and $2.1\times$ speedup. Unpaced, \method{} is 0.6\,s slower, the cost of encoding and replay. With one-step-off pipelining, \method{} runs at about 8\,s per update while full synchronization still takes 11.9\,s and 17.9\,s.

\paragraph{Scaling with worker count.}
\begin{figure}[t]
\centering
\includegraphics[width=\linewidth]{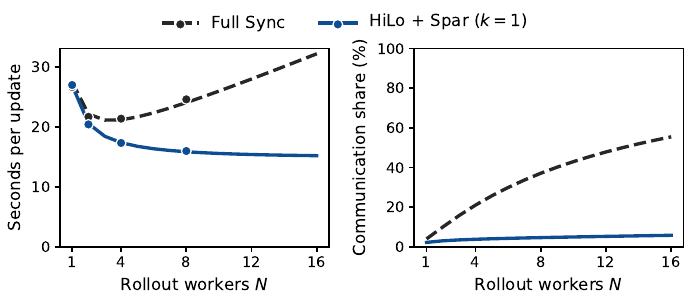}
\caption{Left: lockstep seconds per update as the number of rollout workers $N$ grows behind one learner GPU whose 1\,Gbps uplink is shared by unicast, with the global batch fixed at 16 prompts $\times$ 8 rollouts. Right: share of each update spent on communication. Points are measured (one seed, updates 6--20); lines are the step-time model of this section, fitted to the measured generation time at $N=4$ and $N=8$.}
\label{fig:scaling}
\end{figure}
When workers share the learner's uplink, adding workers shortens generation but multiplies the bytes the learner sends (Figure~\ref{fig:scaling}). Full synchronization is fastest at about four workers (21.4\,s per update) and slows to 24.6\,s at eight, whereas \method{} keeps improving to 16.0\,s, 1.54$\times$ faster. The lockstep model $T=G(N)+L+NS/B$, with generation time $G(N)=g_0+g_1/N$ and learner time $L$, matches every measured point within 1\,s and predicts that communication takes 56\% of a full-synchronization update at $N=16$ against below 6\% for \method{}. By the same model (Appendix~\ref{app:regime}), the gain shrinks with more computation per update, grows with adapter size and workers per link, and vanishes with multi-gigabit links per worker.

\section{Conclusion}
\label{sec:conclusion}
We presented \method{}, which synchronizes LoRA policies in distributed on-policy post-training by sending compressed gradients and letting rollout workers replay the optimizer locally, and \gram{}, a budget-aware backend that decouples training rank from communication rank. \method{} stays close to full synchronization at 1.6\% of the adapter bytes and speeds up each update by up to $2.2\times$ on bandwidth-limited links. More broadly, by decoupling what the learner trains from what it must ship, \method{} makes distributed rollout practical in communication-constrained settings, from geographically distributed clusters to heterogeneous compute shared over wide-area networks, without sacrificing adapter capacity.

\section*{Limitations}
\label{sec:limitations}
\paragraph{Objective coverage.}
The consistency guarantee is objective-independent, but the effect of gradient compression on learning is not, and we measure it only for OPD and RLAD. Pure GRPO gradient-synchronization results are still needed to validate quality beyond these settings.

\paragraph{State replication and reliability.}
Worker state memory depends on the optimizer (\S\ref{sec:optimizer-instances}). For the current rank-64 1.7B configuration ($|\vec{\theta}|=69{,}730{,}304$), the FP32 AdamW moments alone occupy approximately 558\,MB per worker and add to startup and recovery traffic. Exact reconstruction also requires matching optimizer versions, update ordering and numerical kernels; heterogeneous execution, numerical quantization, dropped messages and restarts require empirical validation. State equality does not establish identical sampled trajectories across different inference kernels. The end-to-end links are paced inside one cluster, so wide-area latency, jitter and loss are not modeled.

\bibliography{references}
\appendix
\section{Why the Gradient Compresses Better than the Update}
\label{app:rank}
\setcounter{proposition}{0}
\renewcommand{\theproposition}{\thesection.\arabic{proposition}}
\renewcommand{\theHproposition}{appendix.rank.\arabic{proposition}}
This appendix supports the argument of \S\ref{sec:optimizer-instances}: the gap between \method{} and the Delta baseline, which compresses the post-optimizer parameter change at the same payload with the same error feedback, is created by the optimizer. Under SGD the gradient and the update have identical spectra, momentum raises the update's rank additively, and AdamW's coordinatewise normalization gives the update the rank of the inverse-root-second-moment matrix, which is generically full. The rank statements are exact; the statement about the update's \emph{spectrum} is checked numerically and stated as a measurable prediction. Nothing here is part of the definition of \method{}, which is optimizer-agnostic (\S\ref{sec:replay-consistency}).

\subsection{Notation}
\label{app:rank-notation}
For one adapted matrix $\mat{W}=\mat{W}_0+s\mat{B}\mat{A}$ with $\mat{B}\in\mathbb R^{d_{\rm out}\times r}$ and $\mat{A}\in\mathbb R^{r\times d_{\rm in}}$, let $\vec{x}_\tau\in\mathbb R^{d_{\rm in}}$ be the module input and $\vec{\delta}_\tau=\partial L/\partial(\mat{W}\vec{x}_\tau)$ the backpropagated error for token $\tau$ of a batch of $T$ tokens, stacked as $\mat{X}\in\mathbb R^{T\times d_{\rm in}}$ and $\mat{\Delta}\in\mathbb R^{T\times d_{\rm out}}$. The dense and factor gradients are
\begin{equation*}
\mat{G}=\mat{\Delta}^{\top}\mat{X},\quad
\mat{G}_B=s\,\mat{G}\mat{A}^{\top},\quad
\mat{G}_A=s\,\mat{B}^{\top}\mat{G}.
\end{equation*}
With $\sigma_1(\mat{M})\ge\sigma_2(\mat{M})\ge\cdots$ the singular values, the relative Eckart--Young tail
\begin{equation*}
\tau_k(\mat{M})=\sum_{i>k}\sigma_i(\mat{M})^2\big/\lVert\mat{M}\rVert_F^2
\end{equation*}
is the fraction of $\lVert\mat{M}\rVert_F^2$ that the best rank-$k$ approximation discards; $1-\tau_k$ is the retained energy that Figure~\ref{fig:kk17_energy} reports for exact truncation. $\odot$ and $\oslash$ are entrywise product and division. All statements concern one factor, with the index suppressed.

\subsection{The factor gradient inherits the dense gradient's decay}
\begin{lemma}[spectral inheritance]
\label{lem:inherit}
For every $i$ and $k$,
\begin{equation*}
\begin{aligned}
\sigma_i(\mat{G}_B)&\le s\lVert\mat{A}\rVert_2\,\sigma_i(\mat{G}),\\
\sigma_i(\mat{G}_A)&\le s\lVert\mat{B}\rVert_2\,\sigma_i(\mat{G}),\\
\sigma_i(\mat{G})&\le\sigma_i(\mat{\Delta})\,\sigma_1(\mat{X}),\\
\sigma_i(\mat{G})&\le\sigma_1(\mat{\Delta})\,\sigma_i(\mat{X}),\\
\textstyle\sum_{i>k}\sigma_i(\mat{G}_B)^2&\le s^2\lVert\mat{A}\rVert_2^2\,\textstyle\sum_{i>k}\sigma_i(\mat{G})^2.
\end{aligned}
\end{equation*}
\end{lemma}
\begin{proof}
$\sigma_i(\mat{M}\mat{N})\le\sigma_i(\mat{M})\sigma_1(\mat{N})$ and $\sigma_i(\mat{M}\mat{N})\le\sigma_1(\mat{M})\sigma_i(\mat{N})$ \citep[Thm.~3.3.16]{horn1991topics}, applied to $(\mat{G},\mat{A}^{\top})$, $(\mat{B}^{\top},\mat{G})$ and $(\mat{\Delta}^{\top},\mat{X})$.
\end{proof}
The lemma is a bound, not a guarantee: the factor gradients are never less compressible than $\mat{G}$, and $\mat{G}$ concentrates whenever the per-token errors or activations of a batch occupy few directions, which is the case in OPD and RLAD, where each $\vec{\delta}_\tau$ is a scalar times a direction largely fixed by the teacher or the reward. We make no theoretical claim about this; \citet{zhao2024galore} give conditions under which dense gradients become low-rank, and our premise is empirical (Figure~\ref{fig:kk17_energy}: the rank-1 budget retains 13\% of the error-corrected gradient's energy, the rank-8 budget 66\%). Error feedback flattens $\vec{h}_t$ over training (Appendix~\ref{sec:kk}) but every direction is eventually shipped.

\subsection{The optimizer sets the rank of the update}
\label{app:rank-optimizer}
Fix decoded gradients $\widetilde{\mat{G}}_0,\widetilde{\mat{G}}_1,\ldots$ with $\operatorname{rank}(\widetilde{\mat{G}}_t)\le k$, and let $\mat{\Theta}_{t+1}-\mat{\Theta}_t$ be the parameter change produced by the transition of Equation~\ref{eq:replay}. Under SGD without momentum, $\mat{\Theta}_{t+1}-\mat{\Theta}_t=-\eta_t\widetilde{\mat{G}}_t$ has the same singular vectors and tail as the gradient, so compressing before and after the optimizer discard the same energy; everything below is introduced by the optimizer. For AdamW we write, following Equation~\ref{eq:adamw-theta}, $\mat{\Theta}_{t+1}-\mat{\Theta}_t=-\eta_t\lambda\mat{\Theta}_t-\eta_t\mat{R}_{t+1}$ with $\mat{R}_{t+1}:=\widehat{\mat{m}}_{t+1}\oslash(\sqrt{\widehat{\mat{v}}_{t+1}}+\epsilon)$; the decay term adds a multiple of the full-rank parameters and only makes truncation harder, so the statements concern $\mat{R}_{t+1}$.

\begin{proposition}[momentum]
\label{prop:momentum}
For $\mat{m}_{t+1}=\beta_1\mat{m}_t+(1-\beta_1)\widetilde{\mat{G}}_t$ (Equation~\ref{eq:adamw-m}) or the Muon buffer $\mat{M}_{t+1}=\mu\mat{M}_t+\widetilde{\mat{G}}_t$ of Appendix~\ref{app:optimizer-rules}, $\operatorname{rank}(\mat{m}_{t+1})\le\min((t+1)k,r)$, and the Nesterov input $\widetilde{\mat{G}}_t+\mu\mat{M}_{t+1}$ has rank at most $\min((t+2)k,r)$.
\end{proposition}
\begin{proof}
Subadditivity of rank and induction on $t$; initializing the buffer to the first gradient does not change the bound.
\end{proof}
The bound is attained for gradients in general position (\S\ref{app:rank-witness}); numerically, the effective window $1/(1-\beta_1)\approx10$ governs the momentum's rank.

\begin{proposition}[rank transfer under AdamW]
\label{prop:rank-transfer}
Let $\mat{\Lambda}_{t+1}:=\mathbf 1\oslash(\sqrt{\widehat{\mat{v}}_{t+1}}+\epsilon)$, so $\mat{R}_{t+1}=\widehat{\mat{m}}_{t+1}\odot\mat{\Lambda}_{t+1}$. If $\widehat{\mat{m}}_{t+1}=\vec{u}\vec{v}^{\top}$ has rank one with no zero entries, then $\mat{R}_{t+1}=\operatorname{diag}(\vec{u})\mat{\Lambda}_{t+1}\operatorname{diag}(\vec{v})$ and $\operatorname{rank}(\mat{R}_{t+1})=\operatorname{rank}(\mat{\Lambda}_{t+1})$. In general $\operatorname{rank}(\mat{R}_{t+1})\le\operatorname{rank}(\widehat{\mat{m}}_{t+1})\operatorname{rank}(\mat{\Lambda}_{t+1})$.
\end{proposition}
\begin{proof}
$(\vec{u}\vec{v}^{\top}\odot\mat{\Lambda})_{ij}=u_i\Lambda_{ij}v_j$, and invertible diagonal factors preserve rank; the general case is the Hadamard rank inequality.
\end{proof}
The communication rank $k$ has disappeared: the update has the rank of $\mat{\Lambda}_{t+1}$, which the next proposition shows is generically full, including for second moments that a rank-one gradient stream can produce.

\begin{proposition}[genericity of $\mat{\Lambda}$]
\label{prop:lambda-generic}
Let $\widehat{\mat{v}}\in\mathbb R^{d_{\rm out}\times r}_{>0}$ and $\mat{\Lambda}=\mathbf 1\oslash(\sqrt{\widehat{\mat{v}}}+\epsilon)$.
(a) If $\widehat{\mat{v}}=\vec{a}\vec{b}^{\top}$ and $\epsilon>0$, $\operatorname{rank}(\mat{\Lambda})$ is the smaller of the numbers of distinct entries of $\sqrt{\vec a}$ and of $\sqrt{\vec b}$.
(b) For every $\epsilon\ge0$, $\operatorname{rank}(\mat{\Lambda})<r$ only on a Lebesgue-null set of $\widehat{\mat{v}}$, and the same holds on the set of $\widehat{\mat{v}}=\sum_{i=1}^{\ell}w_i(\vec{a}_i\odot\vec{a}_i)(\vec{b}_i\odot\vec{b}_i)^{\top}$ reachable from $\ell\ge3$ rank-one gradients, as a subset of its own parameter space.
\end{proposition}
\begin{proof}
(a) With $\alpha_i=\sqrt{a_i}$, $\beta_j=\sqrt{b_j}$, $\Lambda_{ij}=\beta_j^{-1}/(\alpha_i+\epsilon/\beta_j)$, so $\mat{\Lambda}$ is a Cauchy matrix with nodes $\alpha_i$ and $\epsilon/\beta_j$ times an invertible diagonal matrix. Rows with equal $\alpha_i$ and columns with equal $\beta_j$ coincide, and on distinct nodes the Cauchy determinant $\prod_{i<i'}(x_{i'}-x_i)\prod_{j<j'}(y_{j'}-y_j)/\prod_{i,j}(x_i+y_j)$ is nonzero.
(b) Every $r\times r$ minor of $\mat{\Lambda}$ is real-analytic in $\widehat{\mat{v}}$ on the connected open set $\mathbb R^{d_{\rm out}\times r}_{>0}$, so it vanishes identically or on a null set; at $\hat v_{ij}=(i+j)^2$ the leading minor is the Cauchy determinant with nodes $i$ and $j+\epsilon$, nonzero for every $\epsilon\ge0$. This witness equals $i^2\cdot1+2\,ij+1\cdot j^2$, a sum of three matrices of the form $(\vec{a}\odot\vec{a})(\vec{b}\odot\vec{b})^{\top}$, so it is reachable from three rank-one gradients, and the minor is real-analytic in $(\vec{a}_i,\vec{b}_i,w_i)$ wherever $\widehat{\mat{v}}$ is entrywise positive, which gives the second claim.
\end{proof}

\begin{corollary}
\label{cor:rank-full}
Let $\epsilon>0$. If every decoded gradient is a positive multiple of one rank-one matrix $\vec{a}\vec{b}^{\top}$ with nonzero entries, $|b_j|$ pairwise distinct and $|a_i|$ taking at least $r$ distinct values, then $\operatorname{rank}(\mat{R}_{t+1})=r$ for every $t\ge0$, although every gradient has rank $k=1$. For rank-one streams of any length, $\operatorname{rank}(\mat{R}_\ell)<r$ only on a null set of the stream parameters.
\end{corollary}
\begin{proof}
For the collinear stream $\widehat{\mat{m}}_{t+1}\propto\vec{a}\vec{b}^{\top}$ and $\widehat{\mat{v}}_{t+1}\propto(\vec{a}\odot\vec{a})(\vec{b}\odot\vec{b})^{\top}$, so Proposition~\ref{prop:lambda-generic}(a) and Proposition~\ref{prop:rank-transfer} give the rank. The leading minor of $\mat{R}_\ell$ is real-analytic in the stream wherever $\widehat{\mat{v}}_\ell$ is entrywise positive, and on every connected component of that domain it is nonzero at a collinear stream perturbed by arbitrarily small later gradients, so it vanishes only on a null set.
\end{proof}

\paragraph{Rank versus spectrum.}
Rank is coarse: a full-rank matrix with one dominant singular value is still well approximated at rank one. If the second moments were uniform, $\mat{\Lambda}_{t+1}\approx c\mathbf 1\mathbf 1^{\top}$ and $\mat{R}_{t+1}\approx c\,\widehat{\mat{m}}_{t+1}$ would be as compressible as the momentum; likewise, with $\epsilon=10^{-8}$ the exact full rank of Corollary~\ref{cor:rank-full} is numerically invisible when $\widehat{\mat{v}}$ is rank one. The practically relevant mechanism is Proposition~\ref{prop:lambda-generic}(b): heterogeneous second moments accumulated over several steps, which is what AdamW's normalization is for. We have no closed-form bound on $\tau_k(\mat{R}_{t+1})$; \S\ref{app:rank-witness} checks the spectrum numerically and \S\ref{app:rank-prediction} states the prediction.

\paragraph{Muon and Lion.}
The Newton--Schulz polynomial of Appendix~\ref{app:optimizer-rules} has the form $\mat{Y}\,p(\mat{Y}^{\top}\mat{Y})$, so the Muon update has rank at most that of its Nesterov input (Proposition~\ref{prop:momentum}), but the orthogonalization drives its nonzero singular values toward a common value, so $\tau_k\approx1-k/\operatorname{rank}$: within its rank it is the least truncatable update possible, consistent with Delta collapsing in every Muon seed (Figure~\ref{fig:countdown_optimizers}). Lion's update $\operatorname{sign}(\mat{c}_{t+1})$ \citep{chen2023lion} is also entrywise and does not preserve low rank beyond the rank-one case, yet Delta matches \method{} + \gram{} under Lion (63.2\% versus 63.7\% pass@4, Appendix~\ref{sec:countdown}). Rank amplification alone therefore does not predict Delta's loss; the magnitude structure of the discarded part matters too, and under Lion every entry has the same magnitude. We leave this case open.

\subsection{Numerical witness}
\label{app:rank-witness}
\begin{table}[t]
\centering
\footnotesize
\setlength{\tabcolsep}{3.4pt}
\renewcommand{\arraystretch}{1.18}
\definecolor{headrow}{gray}{0.96}
\resizebox{\linewidth}{!}{%
\begin{tabular}{c ccc c ccc}
\toprule
& \multicolumn{3}{c}{\textbf{Rank}} & & \multicolumn{3}{c}{\textbf{Retained energy} $1-\tau_1$} \\
\cmidrule(lr){2-4}\cmidrule(l){6-8}
\rowcolor{headrow}{\color{black!70}Update $t$} & $\mat{m}_t$ & $\mat{\Lambda}_t$ & $\mat{R}_t$ & & $\widetilde{\mat{G}}_{t-1}$ & $\widehat{\mat{m}}_t$ & $\mat{R}_t$ \\
\midrule
1 & 1 & 4 & 3 & & 1.00 & 1.00 & 1.00 \\
2 & 2 & 48 & 62 & & 1.00 & 0.72 & 0.59 \\
3 & 3 & 64 & 64 & & 1.00 & 0.49 & 0.37 \\
10 & 10 & 64 & 64 & & 1.00 & 0.66 & 0.20 \\
30 & 30 & 64 & 64 & & 1.00 & 0.20 & 0.26 \\
\bottomrule
\end{tabular}}
\caption{Synthetic AdamW replay of a rank-one gradient stream on one $1024\times64$ factor (\texttt{scripts/appendix\_rank\_witness.py}, seed 0); not a measurement on the trained model. $\operatorname{rank}(\mat{m}_t)=t$ attains Proposition~\ref{prop:momentum}; $\mat{\Lambda}_t$ and $\mat{R}_t$ are numerically full rank from the third update. A rank-one truncation that loses nothing on the gradient discards 60--80\% of the update's energy.}
\label{tab:rank-witness}
\end{table}
Table~\ref{tab:rank-witness} replays AdamW ($\beta_1=0.9$, $\beta_2=0.999$, $\epsilon=10^{-8}$) on a $1024\times64$ factor, the $\mat{B}$ shape of a Qwen3-1.7B attention projection at $r=64$, fed random rank-one gradients $\vec{a}_t\vec{b}_t^{\top}$ with log-normal entry scales so that $\widehat{\mat{v}}$ is heterogeneous; ranks are numerical ranks in double precision. Under SGD with momentum on the same stream, $\operatorname{rank}(\mat{\Theta}_{31}-\mat{\Theta}_{30})=30$, as Proposition~\ref{prop:momentum} alone predicts. The Cauchy witness of Proposition~\ref{prop:lambda-generic}(b) is exact but ill-conditioned (smallest singular value about $10^{-19}$ at $r=64$); it is a proof device, and the random stream is the practically relevant witness.

\subsection{Consequence for Delta and a testable prediction}
\label{app:rank-prediction}
At equal bytes with the same error feedback, the energy discarded before feedback is $\tau_k(\widetilde{\mat{G}}_t)$ for \method{} and $\tau_k(\mat{\Theta}_{t+1}-\mat{\Theta}_t)$ for Delta. The first is controlled by the dense gradient's tail (Lemma~\ref{lem:inherit}), which Figure~\ref{fig:kk17_energy} measures; the second equals the first under SGD and is larger under momentum and AdamW, where the update is generically full rank. \method{} ships the quantity whose tail is set by the data and lets each worker's replicated optimizer regenerate the full-rank update at no communication cost, whereas Delta ships the quantity whose tail is set by the second-moment statistics. This matches Figure~\ref{fig:countdown_optimizers}: Delta is closest to \method{} + \gram{} under SGD with momentum, trails it under AdamW in every non-collapsed seed, and collapses under Muon.

The argument predicts that the compressibility of the Delta message tracks $\mat{\Lambda}_{t+1}$ rather than the gradient. The shadow procedure of Table~\ref{tab:kk_fidelity} computes every quantity involved, so one additional panel of Figure~\ref{fig:kk17_energy} would test it: on the same Full Sync steps, plot $1-\tau_k$ against $k$ for the decoded gradient, the bias-corrected momentum, $\mat{\Lambda}_{t+1}$ and $\mat{R}_{t+1}$. The prediction is that the update's curve follows $\mat{\Lambda}$'s, not the gradient's, and lies below the gradient's at every $k$. We have not run this measurement, so the spectral claim rests on Table~\ref{tab:rank-witness}, and the Lion result shows that rank alone does not determine Delta's loss.

\subsection{Scope}
\label{app:rank-scope}
Everything in \S\ref{app:rank-optimizer} is specific to the optimizer and explains the Delta gap for momentum-based and adaptive updates; it is not a property of the \method{} protocol, whose same-version consistency (\S\ref{sec:replay-consistency}) holds for any replicable transition. Lemma~\ref{lem:inherit} is conditional on the dense gradient's spectrum, for which Figure~\ref{fig:kk17_energy} is the evidence. No statement here bounds $\tau_k(\mat{R}_{t+1})$ analytically, and the Lion result shows that rank alone does not determine Delta's loss; the spectral claim rests on Table~\ref{tab:rank-witness} until the panel described in \S\ref{app:rank-prediction} is measured.

\section{Additional Related Work}
\label{app:related}
\paragraph{Gradient compression.}
PowerSGD \citep{vogels2019powersgd} and Deep Gradient Compression \citep{lin2018dgc} compress gradient communication in data-parallel training, and error feedback compensates omitted directions over time \citep{seide2014onebit,karimireddy2019ef}. 1-bit Adam \citep{tang2021onebitadam} combines communication compression with adaptive optimization through a distinct training recipe, and GaLore \citep{zhao2024galore} gives conditions under which dense weight gradients become low-rank. \method{} does not introduce gradient compression itself. It moves the compressor to a different interface, after aggregation on the learner and before one-to-many dissemination to inference-only workers, where a global byte budget and exact short-side decompositions become practical.

\paragraph{Policy dissemination.}
INTELLECT-2 relays weights through Shardcast \citep{primeintellect2025intellect2}, and ECHO-2 combines bounded staleness, overlap of communication with computation and pipelined peer-to-peer broadcast \citep{song2026echo2}. These mechanisms change how a payload travels rather than what it contains, so \method{}'s gradient messages could be relayed in the same way. Lossless weight-delta codecs shrink the payload itself. Fireworks AI reconstructs the trainer's weights bit for bit from compressed BF16 deltas \citep{fireworks2026arcv3}, and SparseRL-Sync sends only the changed elements of full-parameter updates \citep{hu2026sparserlsync}. Both rely on most weights staying unchanged between versions. A LoRA update is dense in parameter space, so these codecs save little on adapters (\S\ref{sec:systems}), whereas its factor gradients remain compressible. IMPALA \citep{espeholt2018impala} separates actors from learners with off-policy correction, LoRA serving systems \citep{chen2024punica,sheng2024slora} make adapter installation cheap, and infrequent synchronization \citep{douillard2023diloco} trades communication for staleness, which on our Countdown-4 runs costs far more accuracy than gradient compression at the same bytes (Appendix~\ref{app:syncK}). None of them reduces the size of each repeated adapter transfer.

\paragraph{LoGRA.}
Concurrent with this work, LoGRA \citep{zhang2026logra} accumulates random low-rank sketches of the gradients of selected weight matrices during backpropagation and synchronizes rollout engines with the optimizer-adjusted low-rank update. It targets training memory in full-weight RL, whereas \method{} targets learner-to-worker bandwidth for an existing high-rank LoRA adapter. LoGRA runs the optimizer only on the trainer, so rollout engines receive a parameter update of at most the sketch rank, as in our Delta baseline. In \method{}, workers replay the full optimizer transition from the decoded gradient, so an adaptive optimizer such as AdamW can still produce a full-rank update in factor space (Appendix~\ref{app:rank}).

To see whether LoGRA's update carries over to LoRA factors, we apply a LoGRA-style update to the rank-64 Countdown-4 adapter at the \gram{} rank-1 payload (Table~\ref{tab:logra_lora}). Each step sketches every factor gradient along its long side with a fresh seeded random sign matrix, keeps RowAdam state and applies the low-rank update. We do not use LoGRA's predicted-KL step control and sweep the learning rate instead, so this is not a full reproduction of LoGRA. The best setting reaches 51.6\% pass@4 against 64.1\% for \gram{}, and larger learning rates collapse. A random sketch of a factor's long side keeps about 1.6\% of the gradient energy per update here, whereas \gram{} ships the dominant directions exactly and keeps 13.8\%. Random sketches can amortize over many steps in full-weight training, but on a LoRA factor the same payload carries far less signal per update.
\begin{table}[t]
\centering
\footnotesize
\setlength{\tabcolsep}{3.4pt}
\renewcommand{\arraystretch}{1.18}
\providecommand{\std}[1]{\,{\color{black!72}\scriptsize$\pm$#1}}
\definecolor{oursrow}{RGB}{232,240,252}
\definecolor{headrow}{gray}{0.96}
\resizebox{\linewidth}{!}{%
\begin{tabular}{l ccc}
\toprule
& \textbf{pass@4 (\%)} $\uparrow$ & \textbf{mean@4 (\%)} $\uparrow$ & \textbf{Energy (\%)} $\uparrow$ \\
\midrule
LoGRA-style, lr $1{\times}10^{-4}$ & 51.6\std{0.7} & 32.2\std{1.2} & 1.6 \\
LoGRA-style, lr $3{\times}10^{-4}$ & 46.9\std{6.4}$^{1/3}$ & 31.0\std{7.2} & 1.6 \\
LoGRA-style, lr $1{\times}10^{-3}$ & collapsed$^{3/3}$ & -- & -- \\
\rowcolor{oursrow}\textbf{\method{} + \gram{} (ours)} & \textbf{64.1}\std{1.3} & \textbf{51.5}\std{1.2} & \textbf{13.8} \\
\bottomrule
\end{tabular}}
\caption{A LoGRA-style update applied to the rank-64 LoRA factors on Countdown-4 at the \gram{} rank-1 payload (recipe of Table~\ref{tab:main_accuracy}; three seeds per learning rate, five for \gram{}). Each step sketches every factor gradient along its long side with a fresh seeded random sign matrix, keeps RowAdam state and applies the low-rank update; the sketch rank matches the payload (17 for a $1024{\times}64$ factor). Predicted-KL step control is not used, so the learning rate is swept. Energy is the mean fraction of the gradient energy each update retains, in the random sketch for LoGRA and in the shipped directions of the error-corrected gradient for \gram{}. $^{c/n}$: $c$ of $n$ seeds collapsed and are excluded.}
\label{tab:logra_lora}
\end{table}

\paragraph{On-policy objectives and LoRA capacity.}
OPD \citep{agarwal2024gkd}, OPSD \citep{opsd2026}, combinations with RLVR \citep{opdrlvr2026} and reinforcement-aware distillation \citep{zhang2026rlad} differ in supervision and failure modes. Evidence that low-rank RL can retain quality \citep{schulman2025lorawithoutregret} makes Low-rank LoRA an essential baseline.

\section{Additional Experimental Results}
\label{app:results}
\subsection{Knights and Knaves}
\label{sec:kk}
\begin{figure}[t]
\centering
\includegraphics[width=\linewidth]{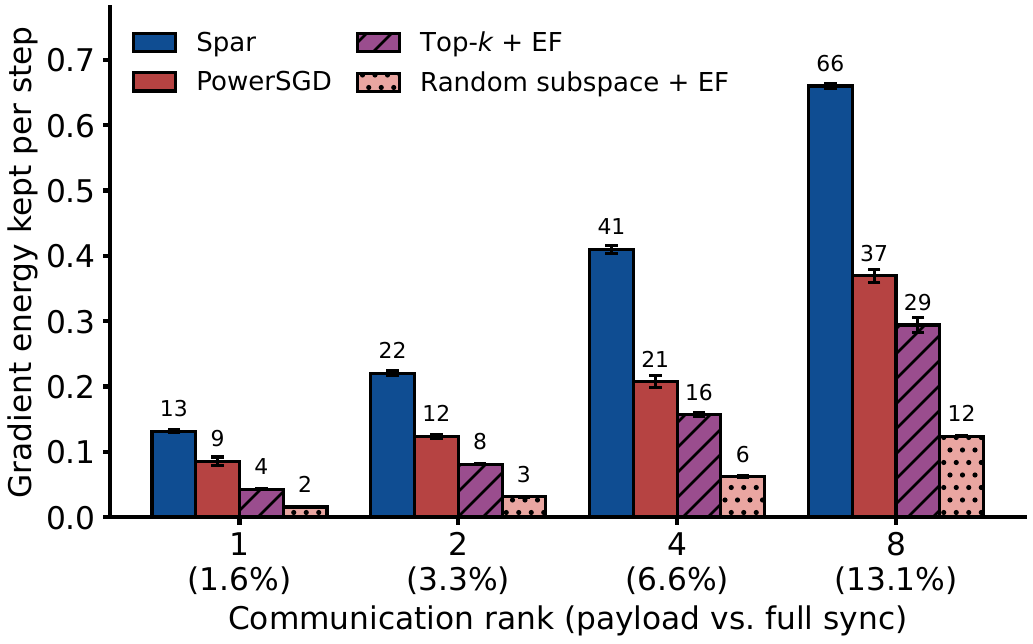}
\caption{Fraction of the error-corrected gradient energy $\lVert\mathbf{h}_t\rVert^2$ that each compressor ships per step, averaged over updates 1--300 of the Qwen3-1.7B KK runs; bars are means over three seeds with sample standard deviations, labels in percent. All compressors send the same number of bytes at each rank.}
\label{fig:kk17_energy}
\end{figure}

\paragraph{Retained gradient energy.}
Figure~\ref{fig:kk17_energy} explains the ordering of the backends in Table~\ref{tab:main_accuracy} and Figure~\ref{fig:kk17_bytes_full}. Per step, \gram{} retains 13\%, 22\%, 41\% and 66\% of the error-corrected gradient energy at ranks 1, 2, 4 and 8, against 9\%, 12\%, 21\% and 37\% for one-iteration PowerSGD. At the same bytes, Top-$k$ spends two thirds of its payload on indices, and a random subspace with error feedback keeps only about $k/64$ of the energy regardless of the spectrum, matching their slower learning. Over training, the retained fraction of every compressor declines as the residual accumulates directions omitted in earlier steps and as the fresh gradient becomes less concentrated once common structure such as the answer format is learned. PowerSGD's single warm-started power iteration falls further behind exact truncation as the spectrum flattens. These are learner-side diagnostics and do not by themselves predict accuracy.

\paragraph{Fidelity of the AdamW update.}
\begin{table}[t]
\centering
\footnotesize
\setlength{\tabcolsep}{3.4pt}
\renewcommand{\arraystretch}{1.18}
\definecolor{oursrow}{RGB}{232,240,252}
\definecolor{headrow}{gray}{0.96}
\begin{tabular}{l cccc}
\toprule
& \multicolumn{4}{c}{\textbf{Relative update error} $\downarrow$} \\
\cmidrule(l){2-5}
\rowcolor{headrow}{\color{black!70}Rank $k$} & $1$ & $2$ & $4$ & $8$ \\
\midrule
Exact, $k$ per factor & 0.558 & 0.511 & 0.438 & 0.338 \\
PowerSGD ($k$ per factor) & 0.582 & 0.549 & 0.493 & 0.402 \\
Exact, AdamW metric & 0.557 & 0.510 & 0.439 & 0.342 \\
\gram{}, AdamW metric & \textbf{0.545} & \textbf{0.498} & \textbf{0.427} & \textbf{0.327} \\
\rowcolor{oursrow}\textbf{\gram{} (ours)} & 0.556 & 0.531 & 0.481 & 0.398 \\
\bottomrule
\end{tabular}
\caption{Relative error $\|U(\hat g)-U(g)\|/\|U(g)\|$ of the AdamW update produced by each compressor's error-corrected gradient, against the update from the full gradient. All compressors run as shadows on the same Full Sync gradients, and only steps containing every listed diagnostic are retained (\numKKfidSteps{} paired steps). \gram{} allocates its budget by raw gradient energy; the AdamW-metric variants allocate by the moment-normalized update instead. Bold marks the lowest error per rank.}
\label{tab:kk_fidelity}
\end{table}

Table~\ref{tab:kk_fidelity} runs every compressor as a shadow on the same uncompressed gradients, each with its own residual history, and measures how far the resulting AdamW update is from the update of the full gradient. Exact truncation with $k$ directions per factor is closer than one-iteration PowerSGD at every tested rank. Global budgeting by raw energy can retain more gradient energy yet give a larger update error at ranks 2 and above, because omitting a small-gradient factor can lose its moment-normalized update. Allocating by an AdamW-aware metric reduces this error, but we do not use it in the training runs.

\subsection{Countdown-4}
\label{sec:countdown}
\paragraph{Seed-level comparison.}
Each seed fixes the initialization and the prompt stream, so Countdown-4 runs can be paired by seed. At rank 1, \gram{} is ahead of PowerSGD and Low-rank LoRA in every seed on both pass@4 and mean@4, by 2.5 and 2.7 pass@4 points on average. Top-$k$ at the same payload is below \gram{} in every seed at every rank, by 3.1, 1.8 and 0.9 pass@4 points at ranks 1, 2 and 4. Delta trails \method{} + \gram{} in 8 of 9 seed--rank pairs. Below one direction per factor, \gram{} reaches 59.7\% pass@4 at 0.39\% of the bytes. Low-rank LoRA has the lowest mean@4 at every rank. With a Qwen3-1.7B student the task becomes rank-insensitive, and the training-reward curves of all rank-1 methods overlap. In the ablation of Table~\ref{tab:spar_ablation}, we reran a collapsed seed of \gram{} without error feedback, and it collapsed again.

\begin{figure}[t]
\centering
\includegraphics[width=\linewidth]{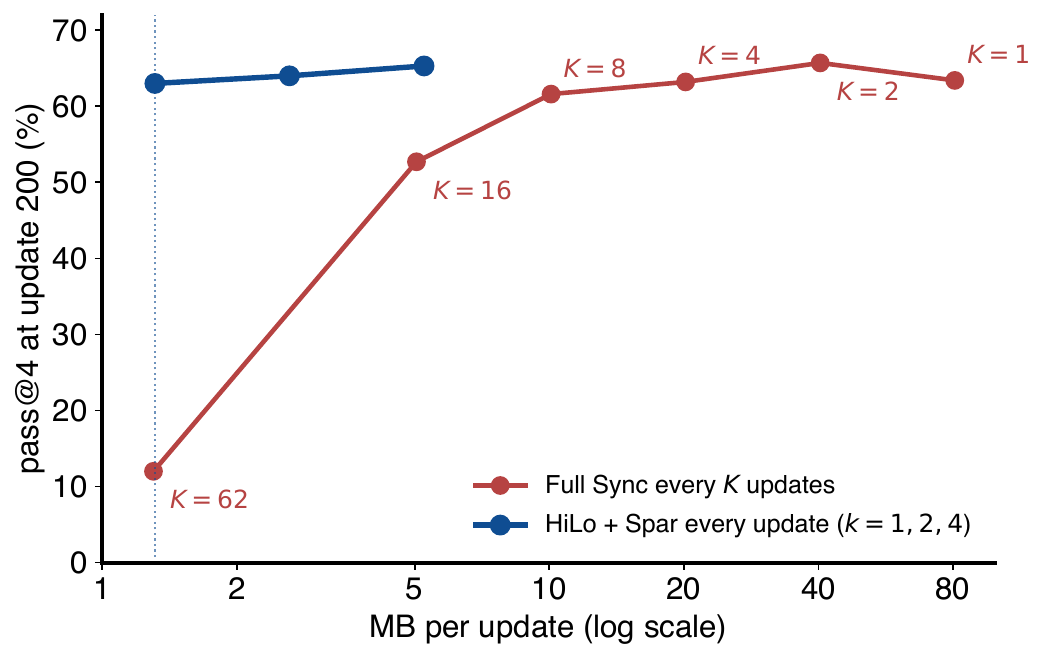}
\caption{Countdown-4 pass@4 at update 200 against bytes per update for Full Sync every $K$ updates and \method{} + \gram{} at every update ($k{=}1,2,4$), with the recipe of Table~\ref{tab:main_accuracy} and one seed. Between full synchronizations, actors keep sampling from the last shipped adapter. The dotted line marks the \method{} rank-1 payload, which Full Sync matches at $K{=}62$.}
\label{fig:countdown_syncK}
\end{figure}

\paragraph{Infrequent full synchronization.}
\label{app:syncK}
Shipping the full adapter less often also cuts traffic, but it raises staleness instead of approximating the gradient. Figure~\ref{fig:countdown_syncK} ships the 80.7\,MB adapter every $K$ updates on Countdown-4, so actors sample from a policy up to $K$ updates old, and the PPO ratio uses their behaviour log-probabilities. Up to $K{=}8$ (10.1\,MB per update) accuracy stays within two pass@4 points of $K{=}1$, but it drops to 52.7\% at $K{=}16$ and to 12.0\% at $K{=}62$, the interval that matches the \method{} rank-1 payload. There the actors lag the learner by 30 updates on average, and the mean actor--learner KL reaches 1.17, against 0.003 for \method{}. \method{} + \gram{} reaches 63.0\%, 64.0\% and 65.3\% at ranks 1, 2 and 4 with one update of lag, so at 5\,MB per update it is 12.6 points above Full Sync every 16 updates. These runs use a single seed and indicate the trend rather than precise gaps.

\paragraph{Cross-optimizer setup.}
\label{app:cross-optimizer}
The comparison in Figure~\ref{fig:countdown_optimizers} tunes the learning rate for Full Sync per optimizer, giving $3\times10^{-4}$ for AdamW, $5\times10^{-4}$ for Muon, $5\times10^{-5}$ for Lion and $0.3$ for SGD with momentum, and shares it across all \method{} backends. Low-rank LoRA diverges under SGD at the shared rate because its rsLoRA scale $\alpha/\sqrt{k}$ is larger, so we report its best rate from $\{0.0047, 0.0375, 0.1\}$. Every arm uses five seeds, and the diverged Muon Full Sync seed is excluded from its reference mean. \method{} + \gram{} reaches 64.1\%, 63.1\%, 63.7\% and 59.7\% pass@4 under AdamW, Muon, Lion and SGD with momentum, against 63.6\%, 65.9\%, 63.1\% and 61.2\% for Full Sync, and mean@4 follows the same pattern. Delta reaches 6.5\% pass@4 under Muon and 9.6\% in its collapsed AdamW seed, where training stops improving within 75 updates and generations either shrink below 150 tokens or run to the length cap. It matches \gram{} under Lion (63.2\% versus 63.7\%) but trails it under SGD with momentum (57.8\% versus 59.7\%) and under AdamW in every non-collapsed seed. Each optimizer is replayed with its own replicated state and numerical implementation, and the online bitwise checks of the replay simulator cover all four.

\section{Training and Evaluation Details}
\label{app:tasks}
\subsection{Baselines and replay simulation}
\label{app:baselines}
PowerSGD and Top-$k$ run as alternative \method{} backends with the same error feedback and payload as \gram{}, so they serve both as competitors and as instances of the framework. Compressor state includes the error-feedback residual, and PowerSGD also keeps its warm start. PowerSGD uses one warm-started power iteration, since additional iterations add computation without reducing the payload. Exact truncation with $k$ directions per factor (``Exact, $k$ per factor'' in Tables~\ref{tab:spar_ablation} and~\ref{tab:kk_fidelity}) is \gram{} without the global budget. Fixed ranks are a baseline configuration choice, and adaptive allocation could also be applied to PowerSGD. Low-rank LoRA ($r{=}k$) instead trains a rank-$k$ adapter and synchronizes it in full. Delta keeps rank 64, runs the optimizer only on the learner, adds the carried error to each factor's parameter change, truncates the result to rank $k$ at the same payload and adopts the truncated parameters itself, so workers need no optimizer state. Delta supports integer ranks only.

In the learning-quality runs, every compressed method makes the learner follow the compressed trajectory. Actors in the \method{} variants obtain their parameters through a replay simulator that decodes each shipped gradient and applies it to a replicated optimizer state, and actors under Delta receive the truncated parameters. These runs therefore isolate how each backend affects learning, and live worker replay is measured separately with \gram{} (\S\ref{sec:systems}). The simulator output is compared bitwise with the learner online in a subset of runs.

\subsection{Knights and Knaves}
Each KK puzzle has six inhabitants, including synonymous truth-teller and liar role pairs, and the scorer requires the final answer block to match the entire oracle assignment. The student is trained with sampled-token reverse-KL OPD to a Qwen3-8B teacher, using AdamW and 16 prompts $\times$ 8 rollouts per update. Actors and evaluation serve the LoRA adapter unmerged through vLLM's LoRA kernels \citep{kwon2023vllm}. Evaluation uses 1,000 questions with no overlap with the training pool and a 2,048-token cap. Checkpoint horizons alter the learning-rate schedule and are not interchangeable. Low-rank LoRA and Top-$k$ at ranks 1 and 2 saved no update-300 checkpoint in their original runs, so we retrained them with the same recipe and seeds for Table~\ref{tab:main_accuracy}.

\subsection{Countdown-4 and the RLAD objective}
\label{app:rlad}
For Countdown-4 we use the trust-region ratio distillation (TRRD) objective of RLAD \citep{zhang2026rlad}. Instead of adding a KL penalty, it replaces the PPO importance ratio by one anchored to a geometric mixture of the behaviour policy $\pi_{\rm old}$ and the teacher $\pi_T$. For token $y_\tau$ of a student rollout with context $c_\tau$, write $\ell^{\vec\theta}_\tau=\log\pi_{\vec\theta}(y_\tau\mid c_\tau)$, and similarly $\ell^{\rm old}_\tau$ and $\ell^{T}_\tau$. Then
\begin{equation}
\begin{aligned}
\log\rho_\tau={}&\alpha\,\bigl(\ell^{\vec\theta}_\tau-\ell^{\rm old}_\tau\bigr)\\
&+(1-\alpha)\,\mathrm{clip}\bigl(\ell^{\vec\theta}_\tau-\ell^{T}_\tau,-1,1\bigr),
\end{aligned}
\label{eq:rlad-ratio}
\end{equation}
and both terms are differentiated with respect to the current student. With $\bar\rho_\tau=\mathrm{clip}(\rho_\tau,1-\epsilon_{\rm low},1+\epsilon_{\rm high})$, the ratio enters the clipped surrogate
\begin{equation}
\mathcal L(\vec\theta)=-\frac{1}{|\mathcal T|}\sum_{\tau\in\mathcal T}\min\bigl(\rho_\tau A_\tau,\ \bar\rho_\tau A_\tau\bigr),
\end{equation}
averaged over all response tokens $\mathcal T$ in the batch. Setting $\alpha{=}1$ recovers the GRPO/PPO ratio and $\alpha{=}0$ anchors the ratio entirely to the teacher. The teacher only scores the student's rollouts and never generates. Advantages are group-normalized as $A=(R-\bar R)/(\mathrm{std}(R)+10^{-4})$ over the 8 rollouts of a prompt and shared by all of its tokens. Because the teacher term is also weighted by the advantage, a group whose rewards are all equal contributes no gradient. We therefore sample 64 prompts per update and keep the first 32 whose groups have non-constant reward. The teacher is Qwen3-8B, and the KL coefficient to the initial policy is zero.

Each Countdown-4 problem gives four numbers and a target, and the student must write an arithmetic expression inside an \texttt{<answer>} tag that uses every number once and evaluates to the target. The reward is 1 for a correct expression, 0.05 for an expression that uses all numbers but misses the target, 0.01 for an invalid expression and 0 without an answer tag. Expressions containing exponentiation or characters other than digits, whitespace, \texttt{+-*/()} and \texttt{.} are scored as invalid without symbolic evaluation, which could otherwise stall the scorer on inputs such as nested powers. Rollouts are sampled at temperature 1 with at most 1{,}536 new tokens. Four actors serve merged BF16 weights and may generate one update behind the learner, and the PPO ratio uses the behaviour log-probabilities they record, which corrects this lag. Evaluation counts only fully correct samples (reward 1), so partial rewards do not enter pass@4 or mean@4.

We follow the original hyperparameters where they apply ($\alpha{=}0.5$, $\epsilon_{\rm low}{=}0.2$, $\epsilon_{\rm high}{=}0.28$ and the $[-1,1]$ clamp of the student--teacher log-ratio). Our setting differs from the original in four respects. We train a rank-64 LoRA adapter with $\alpha_{\rm LoRA}{=}16$ and rsLoRA scaling, use a learning rate of $3\times10^{-4}$ with 10 warmup steps instead of $10^{-6}$, average the loss per token instead of per sequence, and apply the objective to Countdown-4 rather than KK and math. The authors' code was unavailable when we implemented the objective, so our implementation follows the equations of \citet{zhang2026rlad} and has not been checked against the official code. The Countdown-4 results therefore evaluate this objective as a training signal for \method{}, not a reproduction of RLAD. The teacher itself solves only 35.6\% of the held-out problems greedily, so the student's gains come mainly from the reward rather than from imitating the teacher. Because greedy decoding of these students often repeats itself until the length cap, we evaluate four samples per problem at $T{=}1$.

\section{Implementation and System Analysis}
\label{app:optimizer-implementation}
\subsection{Optimizer replay}
\label{app:optimizer-rules}
\paragraph{AdamW.}
For AdamW \citep{loshchilov2019adamw}, $s_t$ contains first/second moments and counters. Its transition includes
\begin{align}
 \vec{m}_{t+1}&=\beta_1\vec{m}_t+(1-\beta_1)\widetilde{\vec{g}}_t,\label{eq:adamw-m}\\
 \vec{v}_{t+1}&=\beta_2\vec{v}_t+(1-\beta_2)\widetilde{\vec{g}}_t^{\odot2},\label{eq:adamw-v}\\
 \vec{\theta}_{t+1}&=(1-\eta_t\lambda)\vec{\theta}_t
 -\eta_t\frac{\widehat{\vec{m}}_{t+1}}{\sqrt{\widehat{\vec{v}}_{t+1}}+\epsilon}.\label{eq:adamw-theta}
\end{align}
Matching bias correction and decay semantics instantiates Equation~\ref{eq:replay}. Coordinatewise normalization does not preserve gradient rank (Appendix~\ref{app:rank}, Propositions~\ref{prop:rank-transfer}--\ref{prop:lambda-generic}); this is a property of the optimizer, not of the \method{} definition.

\paragraph{Muon.}
Muon uses momentum and approximate matrix orthogonalization \citep{jordan2024muon}. For each LoRA factor, let $\mat{G}_t$, $\mat{M}_t$ and $\mat{U}_t$ denote its gradient, momentum and Nesterov input matrices. Our implementation uses $\mat{M}_{t+1}=\mu \mat{M}_t+\widetilde{\mat{G}}_t$ after initialization, a Nesterov input $\mat{U}_t=\widetilde{\mat{G}}_t+\mu \mat{M}_{t+1}$, five FP32 quintic Newton--Schulz steps, and shape-dependent update scaling $0.2\sqrt{\max(a,b)}$. The first momentum buffer is initialized to the first gradient, rather than using an assumed zero-buffer recurrence. Workers must reproduce these details rather than substitute another Muon variant. Scalar rescaling, decay and orientation conventions are part of $\xi_t$; no backward computation is required, but orthogonalization adds replay work.

A quintic step has the form $(a\mat{I}+b\mat{X}\mat{X}^\top+c(\mat{X}\mat{X}^\top)^2)\mat{X}$, so in exact arithmetic it cannot increase the input matrix's rank. Momentum raises the rank additively (Proposition~\ref{prop:momentum}). Thus Muon and AdamW can satisfy the same synchronization condition while responding differently to compression. Training capacity $r$ is decoupled from communication rank $k$, without promising identical high-rank updates across optimizers.

\paragraph{Other transitions.}
SGD with momentum and Lion \citep{chen2023lion} are replay-eligible when their state, shared hyperparameters and numerical execution are reproduced, and \S\ref{sec:cross-optimizer} evaluates them together with AdamW and Muon. Any extra information an optimizer requires, such as curvature estimates or fresh randomness, is not implicitly available to inference workers and must be reproduced or transmitted.

\paragraph{The wire representation is part of the optimizer.}
If factors are transmitted in BF16 but the learner uses a pre-quantization FP32 reconstruction, the consistency argument of \S\ref{sec:replay-consistency} does not apply. A replay implementation must make the learner consume the same quantized, decoded gradient as workers. Even identical factors can produce slightly different decoded matrices across GPU architectures or GEMM implementations. Matching parameter/state dtypes, tensor ordering, optimizer implementation and update configuration is therefore required for bitwise equality. For AdamW this includes bias correction and fused/foreach execution; for Muon it includes orthogonalization coefficients, iteration count, precision and GEMM execution. Randomized transitions also require matching random draws or counter-based seeds. A heterogeneous implementation must instead measure deviations rather than assume equality.

\paragraph{Replay check procedure.}
\label{app:replay-check}
The learner (Qwen3-0.6B-Base, rank-32 LoRA, budgeted Gram compression at budget 1, two-GPU data parallel) rounds every shipped gradient factor to BF16 and consumes exactly the decoded gradient, as same-version consistency requires (\S\ref{sec:replay-consistency}). It records the initial LoRA parameters and each step's BF16 factors. A separate process on a different GPU reads only these inputs, rebuilds each factor gradient with the learner's batched reconstruction, and runs AdamW with the same hyperparameters and learning-rate schedule. Parameters are compared at every tenth update over 60 updates of KK distillation, and replaying the 60 updates takes 26\,s including loading. Both sides use GH200 GPUs and the same software stack; heterogeneous hardware would require measuring, rather than assuming, equality.

\subsection{Message ordering and recovery}
\label{app:protocol-details}
Each message identifies its optimizer epoch, predecessor and target versions, tensor layout, factor ranks and dtypes, and shared configuration $\xi_t$, including scheduler index and implementation identity. Workers execute updates exactly once in predecessor order, using acknowledgments and duplicate suppression; state checksums can detect divergence. Lockstep rollout collection waits for a committed version. Asynchronous workers may use an older committed version, but must consume every intervening update. Replacing a backlog with only the latest gradient is generally invalid for a stateful optimizer. Updates may be aggregated only when equivalence is established for the chosen transition. Recovery requires a retained update log or a snapshot containing both parameters and optimizer state; an adapter-only restart cannot recover omitted historical state. Backlog storage and recovery traffic contribute to system costs.

\subsection{\gram{} implementation}
\label{app:evd}
\paragraph{Orientation and cost.}
For $a_j<b_j$, \gram{} uses $\mat{H}_j\mat{H}_j^\top$ and its leading eigenvectors $\mat{U}_{j,k_j}$, with $\mat{P}_j=\mat{U}_{j,k_j}$ and $\mat{Q}_j=\mat{H}_j^\top\mat{U}_{j,k_j}$. Both orientations reconstruct $\mat{P}_j\mat{Q}_j^\top$, and a zero-rank factor decodes to zero. Before finite-precision effects this is the best rank-$k_j$ approximation in Frobenius norm, i.e.\ it minimizes the tail $\tau_{k_j}(\mat{H}_j)$ of Appendix~\ref{app:rank-notation}. The decoded output can differ from the ideal truncation because of BF16 quantization and reconstruction arithmetic, which is why both sides consume the same canonical decoder output. For $d\times r$ factors, encoding costs $O(dr^2+r^3+drk_j)$ for Gram construction, eigendecomposition and factor formation, and shape bucketing batches these operations across tensors. We accumulate and decompose Gram matrices in FP64 to reduce the loss of small directions in ill-conditioned residuals; this improves robustness without eliminating conditioning or wire-quantization error.

\paragraph{Greedy allocation.}
Algorithm~\ref{alg:spar-greedy} gives the greedy allocation heuristic. The allocator stably sorts individual directions by descending $\sigma_{j,\ell}^2/c_j$, resolving ties by fixed tensor and direction order. It retains the longest prefix of this global ordering whose cumulative cost fits the budget, stopping at the first direction that would exceed it. Later directions are not considered, even if a cheaper direction would fit the remaining budget, so some feasible payload capacity can remain unused. Within each factor, equal direction costs and non-increasing singular values ensure a spectral prefix. Unequal direction costs make Equation~\ref{eq:budget} a discrete allocation problem, and the greedy rule is not globally optimal. Headers, tensor identifiers and recovery traffic are excluded from the analytical factor budget and must be accounted for in actual transport.

\begin{algorithm}[!ht]
\caption{Greedy rank allocation for \gram{}}
\label{alg:spar-greedy}
\small
\begin{algorithmic}[1]
\Require Descending spectra $\{\sigma_{j,\ell}\}_{\ell=1}^{m_j}$ for $j=1,\ldots,J$
\Require Shapes $(a_j,b_j)$, $m_j=\min(a_j,b_j)$; $q>0$, $\mathcal B\geq0$
\Ensure Ranks $\{k_j\}_{j=1}^J$ and used factor-payload bytes $u$
\State $c_j\gets q(a_j+b_j)$, $k_j\gets0$ for all $j$; $u\gets0$
\State List all $(j,\ell)$ in increasing $j$, then increasing $\ell$
\State Stably sort this list by decreasing $\sigma_{j,\ell}^2/c_j$
\For{each $(j,\ell)$ in the sorted list}
  \If{$u+c_j>\mathcal B$}
    \State \textbf{break}
  \EndIf
  \State $k_j\gets k_j+1$; $u\gets u+c_j$
\EndFor
\State \Return $\{k_j\}_{j=1}^J$, $u$
\end{algorithmic}
\end{algorithm}

\paragraph{Eigensolver.}
The encoder timings in \S\ref{sec:systems} cover error feedback, Gram products, batched FP64 Jacobi eigendecompositions of the $64\times64$ Grams and the global budget, with medians over 30 updates. Two-iteration PowerSGD takes 48--61\,ms on the same gradients. \gram{} thus computes the exact truncation faster than one power iteration. The Jacobi kernel matters because \texttt{torch.linalg.eigh} batches on cuSOLVER only up to $n=32$ and otherwise loops over the matrices, which takes 300\,ms for the same Grams. The end-to-end runs still used that path, so their \method{} times include about 0.27\,s per update that the Jacobi encoder removes; this is an estimate from the encoder measurements, not a rerun.

\subsection{Communication and state accounting}
\label{app:model}
A BF16 adapter with $|\vec{\theta}|$ trainable entries requires $2|\vec{\theta}|$ parameter bytes. Each replaying worker additionally holds $8|\vec{\theta}|$ bytes of FP32 AdamW moments or $4|\vec{\theta}|$ bytes of FP32 Muon momentum, plus $4|\vec{\theta}|$ bytes of FP32 master parameters when used, and serving copies and decode buffers come on top. Factor messages reduce steady-state traffic but do not remove the initial state transfer or recovery snapshots.

\paragraph{Lockstep cost model.}
With $N$ workers sharing a learner uplink of bandwidth $B$ by unicast, generation time $G(N)$, learner time $L$ and message size $S$, the lockstep model of \S\ref{sec:systems} gives $T_{\rm full}\approx G(N)+L+NS_{\rm full}/B$ and
\begin{equation*}
T_{\rm replay}\approx G(N)+L+T_E+NS_{\rm grad}/B+T_D+T_\Phi,
\end{equation*}
where $T_E$, $T_D$ and $T_\Phi$ are the encoding, decoding and worker replay times on the critical path. Compression pays off only if its transfer saving exceeds these added costs, and if $G(N)+L$ dominates both transfer terms, even a large payload reduction gives little step-time improvement. For asynchronous workers, a worker at version $v$ can match the learner's recorded state at $v$ while the learner has moved on, so reconstruction error should be reported separately from version lag.

\paragraph{When synchronization is a bottleneck.}
\label{app:regime}
Applying the model of \S\ref{sec:systems}, which matches the measured points within 1\,s, beyond the measured range shows where the saving matters. At 100\,Mbps per worker, \method{} at rank 1 is 1.7$\times$ faster than full synchronization for the rank-64 Qwen3-1.7B adapter at 15\,s of computation per update, 1.2$\times$ at 60\,s and 1.04$\times$ at 240\,s. For a rank-64 Qwen3-8B adapter (349\,MB, computed from the layer shapes) the same settings give 2.7$\times$, 1.4$\times$ and 1.1$\times$. Longer generation therefore reduces the lockstep benefit at a fixed number of workers, while larger adapters and more workers per link increase it. Asynchrony does not remove the cost either. To hide a 349\,MB transfer at 100\,Mbps behind a 7.8\,s learner update, workers must run four updates behind the learner, whereas \method{}'s factors fit within one. Whether such staleness is acceptable depends on the objective, and we leave its learning cost to future work.

\end{document}